\documentclass[letterpaper, 10 pt, conference]{ieeeconf}  % Comment this line out if you need a4paper

\IEEEoverridecommandlockouts                              % This command is only needed if 
\usepackage{epstopdf}
\usepackage{amsmath,amssymb,xspace,epsfig}
\usepackage{enumerate}
\usepackage{hyperref, url}
\usepackage{multirow}
\usepackage[linesnumbered,ruled]{algorithm2e}
\usepackage{cite}
\usepackage{subfigure}

\def\etal{\emph{et~al.}\xspace}

\newtheorem{definition}{Definition}

\newcommand{\denselist}{\itemsep 0pt\parsep=0.8pt\partopsep 0pt}
\newcommand{\bitem}{\begin{itemize}\denselist}
\newcommand{\eitem}{\end{itemize}}
\newcommand{\benum}{\begin{enumerate}\denselist}
\newcommand{\eenum}{\end{enumerate}}

\begin{document}

\title{Robot Coverage Path Planning for General Surfaces Using Quadratic Differentials 
%\thanks{J. Gao and Chien-Chun Ni is partially supported by grants from AFOSR (FA9550-14-1-0193) and NSF (DMS-1418255, DMS-1221339, CNS-1217823).}
}

\author{
Yu-Yao Lin$^{1}$,
Chien-Chun Ni$^{1}$,
Na Lei$^{2}$,
Xianfeng David Gu$^{1}$ and 
Jie Gao$^{1}$ 
% \thanks{*This work was not supported by any organization}% <-this % stops a space
\thanks{$^{1}$Department of Computer Science, Stony Brook University, Stony Brook, NY, USA
        {\tt\small \{yuylin,chni,jgao,gu\}@cs.stonybrook.edu}}%
\thanks{$^{2}$School of Software, Dalian University of Technology, Liaoning, China
        {\tt\small nalei@dlut.edu.cn}}%
}

\maketitle
%abstract

\begin{abstract}
Robot Coverage Path planning (i.e., provide full coverage of a given domain by one or multiple robots) is a classical problem in the field of robotics and motion planning. The goal is to provide nearly full coverage while also minimize duplicately visited area. In this paper we focus on the scenario of path planning on general surfaces including planar domains with complex topology, complex terrain or general surface in 3D space. The main idea is to adopt a natural, intrinsic and global parametrization of the surface for robot path planning, namely the \emph{holomorphic quadratic differentials}. Except for a small number of zero points (singularities), each point on the surface is given a $uv$-coordinates naturally represented by a complex number. We show that natural, efficient robot paths can be obtained by using such coordinate systems. The method is based on intrinsic geometry and thus can be adapted to general surface exploration in 3D. 

\end{abstract}

%introduction 

\section{Introduction}

The Coverage Path Planning (CPP) problem is to determine a path that passes through all points in a given geometric domain. It is a classical problem in robotics and motion planning and is of fundamental value to many applications that require a robot or multiple robots to sweep over the target area, such as vacuum cleaning robots, lawn mowers, underwater imaging/scanning robots, window cleaners, and many others.

In general the coverage path planning problem has multiple goals: full coverage (i.e., every point in the domain $\Omega$ is covered), no overlapping or repetition (no point is visited multiple times), and/or a variety of objectives on the simplicity or quality of the paths. Satisfying all such requirements is difficult if not impossible. Therefore priorities are often set on these possibly conflicting objectives and the goal is to obtain a good tradeoff. 

The geometric shape of the domain to be covered is crucial in the design of coverage path planning algorithms. Simple shapes such as convex polygons can be covered by simple zigzag motion patterns (lawn mower patterns). Therefore, most algoritms for coverage path planning first decompose the target region into `simple cells'. The cell decomposition can be represented by a cell adjacency graph in which each cell is a vertex and two vertices are connected if they share common boundaries. Within each cell we can use a simple zig-zag pattern and to cover the entire domain we need to visit each cell at least once. 

%Then a tour is obtained by first finding an `exhaustive path' that visits each node of the graph exactly once.  Then this walk is used to generate a coverage path within each cell.
 
In all the decomposition methods, there are two general issues that may affect the final performance. First we need to find a path on the cell adjacency graph that visits each cell at least once -- ideally exactly once (to keep the path short). Finding a path that visits each vertex of a graph exactly once is the well known Hamiltonian path problem, which is NP-hard~\cite{Garey1990computers}. The adjacency graph may not admit a Hamiltonian path -- thus a robot may have to repeatedly visit some points just to get from one cell to the next cell. Second, all the algorithms above use the \emph{extrinsic coordinate system}, i.e., the Euclidean coordinates representing the domain of interest. Such extrinsic coordinate systems, albeit being natural choices, are not the best to encode the complex geometric and topological features introduced by obstacles and boundaries. This is in fact the core challenge that the cell decomposition is mean to tackle. When the domain is not flat (e.g., on a terrain or as a general surface in 3D), the extrinsic coordinate system and the cell decomposition may lead to unnecessarily many pieces depending on the detailed implementation. 

\begin{figure}[t]
	\vspace*{4pt}
	\centerline{\includegraphics[width=0.5\columnwidth]{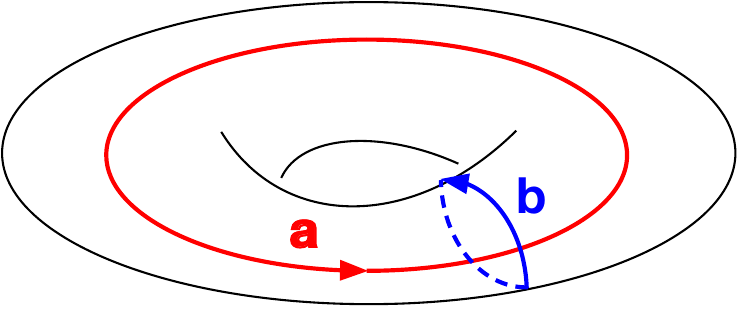}\includegraphics[width=0.5\columnwidth]{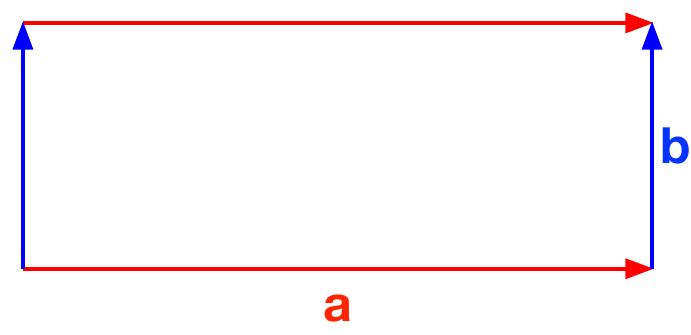}}
	\caption{The torus is sliced open along the two generators, $a$ and $b$, of the fundamental group of the torus.}\label{fig:torus}
	\vspace*{-6pt}
\end{figure}

\medskip\noindent\textbf{Our Contribution.}
In this paper we focus on solving this problem using a generic solution that is applicable to general surfaces in 3D. The novelty of our method is to abandon the extrinsic Euclidean coordinates system and adopt the \emph{intrinsic coordinate system}, i.e., a global parametrization of the domain of interest. To get an idea, consider a standard torus, one can slice the torus open along the two generators of the fundamental group of the torus (See Figure~\ref{fig:torus} as an example) and the torus can be flattened as a square. Thus, one can represent the points on the torus by a $uv$ coordinate system, where the $u$ coordinate represents the position of the point $p$ along one generator of the fundamental group and $v$ represents the position along the other generator. Both the geometry of the surface and the topology of the surface are inherently encoded in this new coordinate system. Finding a coverage path for the torus under the $uv$ coordinate system is now trivial -- one can simply zig-zag in the $uv$ coordinate system which becomes a spiral motion on the torus. 

% \textbf{Please add an example of torus and the two fundamental groups.}

\begin{figure}[htbp]
	\centering
	% \vspace{-6mm}
	\subfigure[By Trapezoid decomposition]{
		\label{fig:road3:subfig:trap}
		\includegraphics[width=0.7\columnwidth]{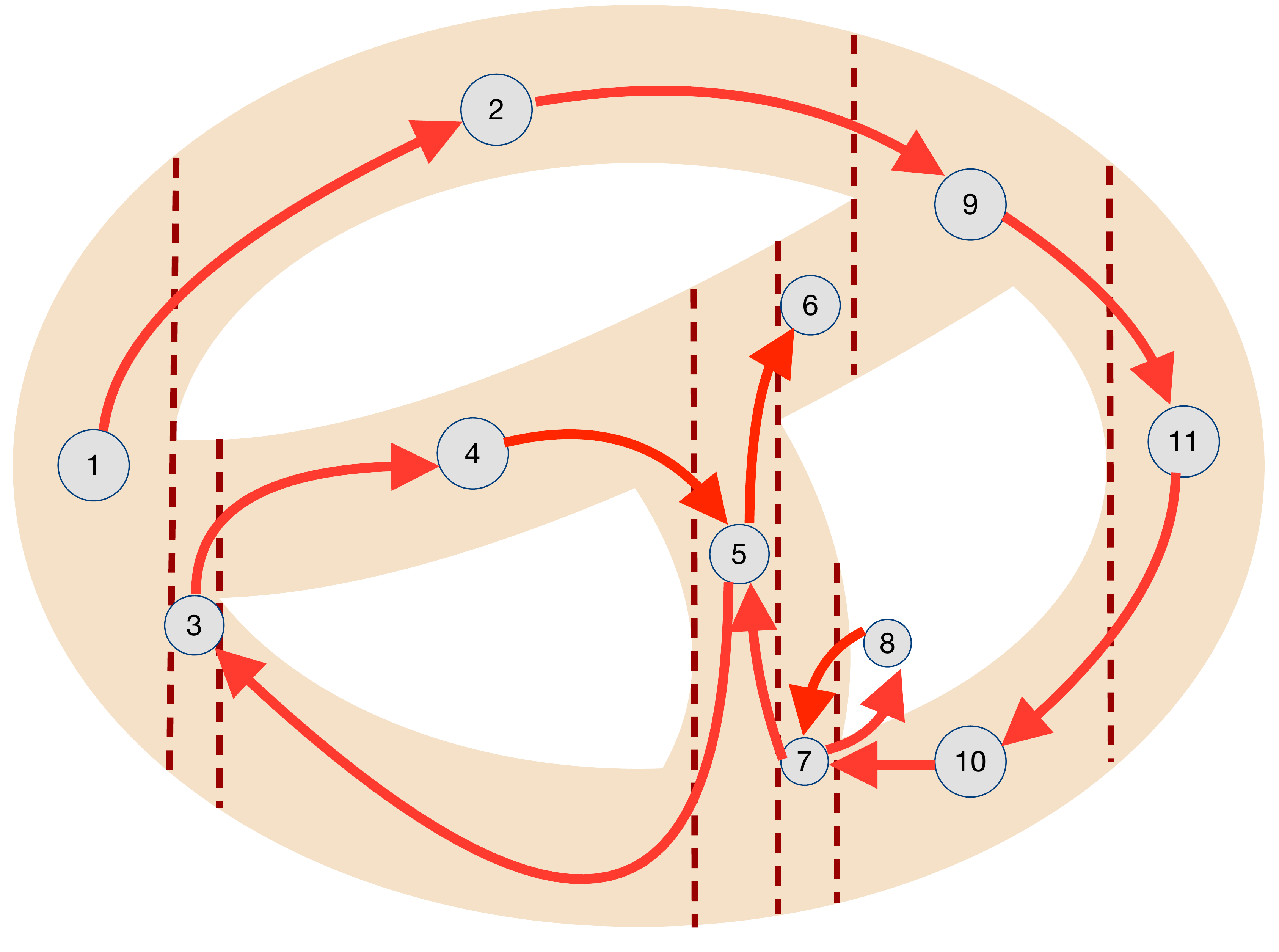}}
	% \hspace{0.0in}
	\subfigure[By Holomorphic Quadratic Differentials]{
		\label{fig:road3:subfig:2form}
		\includegraphics[width=0.7\columnwidth]{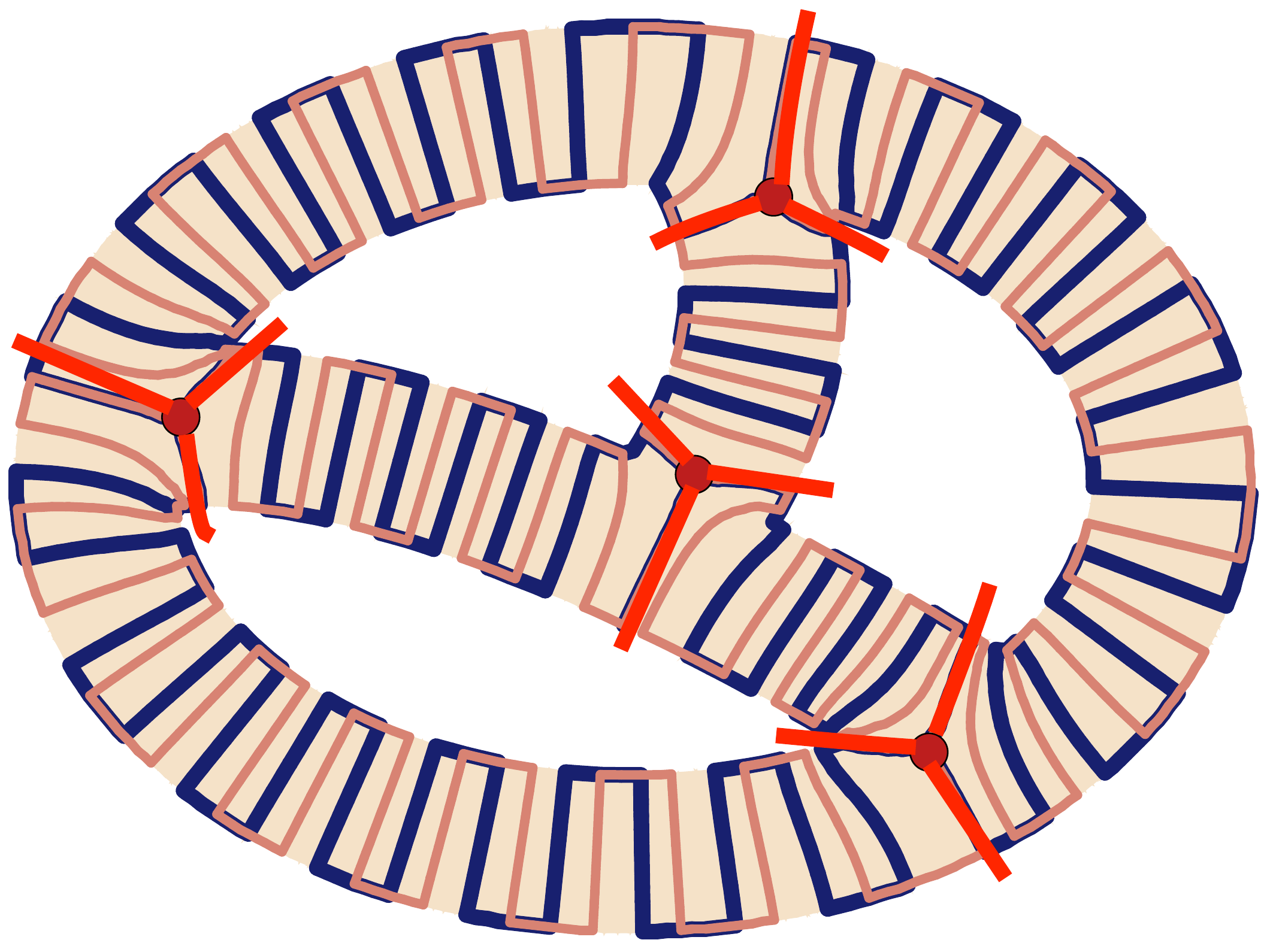}}
	% \vspace{-2mm}
	\caption{Example of a three holes donut with trapezoid decomposition(\ref{fig:road3:subfig:trap}) and our holomorphic quadratic differentials method(\ref{fig:road3:subfig:2form}). In trapezoid decomposition, the donut is decomposed into $11$ cells, the CPP problem here is equivalent to find a Hamiltonian path with these cells as vertices, which is NP-hard. Instead, our method simply cuts the donut into $6$ cells, and the CPP problem in our setting is equivalent to finding Euler cycle with these cells as edges, which can be easily achieved in polynomial complexity.}
	% \vspace{-4mm}
	\label{fig:road3:subfig} %% label for entire figure
\end{figure}

We introduce the theory and algorithms for computing the intrinsic coordinate system using \emph{holomorphic quadratic differentials}. Depending on the topology of the surface there are a constant number of \emph{zero points} (also called critical points, singular points or singularities) which do not have such coordinates. But such singular points are of zero measure. The coordinate system is naturally represented by a complex number. One can trace out a curve by fixing the real/imaginary part of the coordinate, called the vertical/horizontal trajectory respectively. This coordinate system naturally produces a space decomposition by slicing along critical trajectories (i.e., trajectories that end at zero points). Each component is a simply connected piece with the complex coordinates as its natural parametrization. This decomposition can also be represented by a graph $G$ in which the vertices are the critical points and an edge represents a cell that touches two singular vertices. This graph and the coordinate system/parametrization are used to generate a coverage path. See Figure~\ref{fig:road3:subfig} for an example.

To generate a coverage path, we need to decide what order we use to visit the decomposed cells. Again we encounter the problem of visiting each cell at least once. In our setting we actually need to visit all the edges (representing the cells) in $G$, ideally once and only once. So this is in fact the Euler cycle problem, which is, fortunately, much easier than Hamiltonian cycle problem.  Any graph in which all vertices have even degree has an Euler cycle. In our case, the degree of a critical point may not be even. But we can simply double each edge in the graph to create a graph $G'$ (which satisfies the requirement) and compute an Euler cycle on $G'$ -- equivalently, each edge in $G$ is visited precisely twice. This means each cell in the decomposition is visited exactly twice and we can simply stretch and shift the zig-zag pattern in the cell such that the paths followed by the two separate visits have minimum overlap. 

%\textbf{Please add references to Figure 2, 3, 4 to help with the above description. Figure 3 and 4 can be put side by side to save space. Figure 6: need to mark which is $p_1$.}

The main theoretical contribution of this paper lies in the new discrete algorithm for computing holomorphic quadratic differentials for parametrizing the input domain. In the literature, a special subset of holomorphic quadratic differentials, named the 
holomorphic differentials (holomorphic one-forms), have been widely used in computer graphics for surface registration and texture mapping~\cite{gu2002computing,gu2003global}. Compared to this limited subset, holomorphic quadratic differentials construct a larger family of surface parameterization with more freedom and flexibility. The algorithm presented here for holomorphic quadratic differentials is new and has never been published before. Mathematically, holomorphic quadratic differentials are obtained by multiplying two holomorphic one-forms, and their parameterizations should satisfy the property of being a curl-free vector field. It is a challenging problem to control the numerical error around critical points due to the special local structure. Fortunately our robot cover path avoids the zero points and all we need is to trace out the 3 critical trajectories through each critical point, which is carefully handled in our algorithm. 

We evaluated the coverage path generated by our algorithms on a variety of different settings including flat domains with obstacles, non-flat terrains, as well as general high genus surfaces. Our method is an offline method and requires the domain to be known in advance.

\section{Theory on Holomorphic Quadratic Differentials}
\label{sec:prelim}
Our solution for the CPP problem is based on a global surface parameterization, namely the \emph{holomorphic quadratic differentials}. Holomorphic quadratic differentials possess a good property that they inherently induce non-intersecting trajectories on a surface. This benefit prompts us to develop a path planning algorithm based on the trajectories.

Holomorphic quadratic differentials form a branch of study in complex manifold. In this section, we briefly introduce some basics of holomorphic quadratic differentials. Then we design our path planning method on general surfaces. For detailed treatments, we refer readers to~\cite{farkas1992riemann} for Riemann surface theory,~\cite{nehari1975conformal} for complex analysis, and~\cite{strebel1984quadratic} for holomorphic quadratic differentials.

\subsection{Riemann Surfaces} % (fold)
\label{sub:riemann_surfaces}
\begin{definition}(Manifold).
	Let $M$ be a topological space. For each point $p\in M$, there is a neighborhood $U_{\alpha}$ and a continuous bijective map $\phi_{\alpha}:U_{\alpha}\rightarrow V_{\alpha}$ from $U_{\alpha}$ to an open set $V_{\alpha}\subset \mathbb{R}^n$. $(U_{\alpha},\phi_{\alpha})$ is called a local chart. If two neighborhoods $U_{\alpha}$ and $U_{\beta}$ intersect, then the transition map between the chart
	\begin{equation*}
		\phi_{\alpha\beta}=\phi_{\beta}\phi_{\alpha}^{-1}:\phi_{\alpha}(U_{\alpha}\cap U_{\beta})\rightarrow\phi_{\beta}(U_{\beta}\cap U_{\alpha})
	\end{equation*}
	is a continuous bijective map. $M$ is an $n$ dimensional manifold, the set of all local charts $\{(U_{\alpha},\phi_{\alpha})\}$ form an atlas.
\end{definition}
% subsection riemann_surfaces (end)
% Intuitively, a holomorphic differential is a tangent vector field which induces a conformal parameterization. which is an angle-prserving map.

\begin{definition}(Holomorphic Function). 
	A complex function $f :\mathbb{C}\rightarrow\mathbb{C}: x+iy \mapsto u(x,y)+iv(x,y)$ is \emph{holomorphic}, if it satisfies the following Cauchy-Riemann equation
\begin{equation*}
\frac{\partial u}{\partial x} = \frac{\partial v}{\partial y}, \,
\frac{\partial u}{\partial y} = -\frac{\partial v}{\partial x}.
\end{equation*}
\end{definition}
If $f$ is invertible and $f^{-1}$ is also holomorphic, then $f$ is called a \emph{bi-holomorphic} function.

\begin{definition}(Riemann Surface). 
	A Riemann surface is a surface with an atlas $\{(U_{\alpha},\phi_{\alpha})\}$, such that all chart transitions $\phi_{\alpha\beta}$ are bi-holomorphic. The atlas is called a conformal atlas and the local coordinates $\phi_{\alpha}(U_{\alpha})$ are called holomorphic coordinates. The maximal conformal atlas is called a conformal structure of the surface.
\end{definition}

On a Riemann surface, we can define a differential based on the conformal structure. Intuitively, a differential can be regarded as a vector field on a surface. The integration on a differential gives a surface parameterization. The holomorphic differentials and quadratic differentials we introduce below are curl and divergence free vector fields.

\begin{definition}(Holomorphic Differential).
	Given a Riemann surface $R$ with a conformal atlas $\{(U_{\alpha},\phi_{\alpha})\}$, a holomorphic differential $\zeta$ is a complex differential form defined by a family $(U_{\alpha},z_{\alpha},\zeta_{\alpha})$, such that $\zeta_{\alpha}=\phi_{\alpha}(z_{\alpha})dz_{\alpha}$, where $\phi_{\alpha}$ is a holomorphic function on $U_{\alpha}$, and if $z_{\alpha}=\phi_{\alpha\beta}(z_{\beta})$ is the coordinate transformation on $U_{\alpha}\cap U_{\beta}$, then $\phi_{\alpha}(z_{\alpha})\frac{dz_{\alpha}}{dz_{\beta}}=\phi_{\beta}(z_{\beta})$.
% the local representation of the differential form $\zeta$ satisfies the chain rule.
\end{definition}

According to the Poincar{\'e}-Hopf theorem~\cite{hazewinkel2001poincare}, any vector field on a surface with non-zero Euler number must have the singularities where the vector field vanishes. Such singularities are called \emph{zero points}. Here we define the zero points of a holomorphic differential.
\begin{definition}(Zero Point).
	For a point $p$ on a surface $R$, if the local representation of a holomorphic differential $\zeta$ around $p$ is $\zeta_{\alpha}=\phi_{\alpha}(z_{\alpha})dz_{\alpha}$ and $\phi_{\alpha}=0$ at $p$, then $p$ is called a \emph{zero points} of $\zeta$.
\end{definition}

% All oriented two dimensional manifolds are Riemann surfaces. The cases considered in this work, such as 2D or 3D terrains, are also Riemann surfaces. We simply term a Riemann surface as a surface.
\subsection{Holomorphic Quadratic Differentials} % (fold)
\label{sub:quadratic_differential}
\begin{definition}(Holomorphic Quadratic Differential).
	Given a Riemann surface $R$. Let $\Phi$ be a complex differential form with a conformal atlas $\{(U_{\alpha},\phi_{\alpha})\}$, such that on each local chart with the local parameter $z_{\alpha}$,
	\begin{equation*}
		\Phi_{\alpha}=\phi_{\alpha}(z_{\alpha})dz_{\alpha}^{2},
	\end{equation*} 
	where $\phi_{\alpha}(z_{\alpha})$ is a holomorphic function.
\end{definition}

\subsubsection{Zero Points and Trajectories} % (fold)
\label{ssub:zero}

For a holomorphic quadratic differential $\Phi$ on a surface $R$, any point $p\in R$ away from zero has the local coordinate defined as
\begin{equation}
\label{eq:natural}
	\xi(p):=\int^{p}\sqrt{\phi_{\alpha}(z_{\alpha})}dz_{\alpha}.
\end{equation}
This is called the \emph{natural coordinate} induced by $\Phi$. The curves with constant real natural coordinates are called the \emph{vertical trajectories}; while the curves with constant imaginary natural coordinates are called the \emph{horizontal trajectories}. A trajectory which ends in zero points is called a \emph{critical trajectories}, otherwise it is a \emph{regular trajectory}. The horizontal trajectories of $\Phi$ are either infinite spirals or finite closed loops. This means that the trajectories of holomorphic quadratic differentials are non-intersecting trajectories on a surface. This property is the key idea of our path planning algorithm.

\begin{definition}(Genus).
	A genus $g$ of a surface is the largest number of cuttings along non-intersecting simple closed curves on the surface without disconnecting it.
\end{definition}

The local structure around a zero point of a holomorphic quadratic differential is a complex function $z\rightarrow z^{\frac{3}{2}}$. For any holomorphic quadratic differential $\Phi$ on a closed surface with genus $g>1$, there are $4g-4$ zero points. For a multiply-connected surface with $n>2$ boundaries, there are $2g-2$ zero points of $\Phi$. Zero points are also called \emph{critical points} because they are the endpoints of critical trajectories.

\subsection{Surface Decomposition} % (fold)
\label{sub:domain_decomposition}

The path planning technique proposed in this work is applicable to both multiply-connected surfaces(surface with boundaries or obstacles) and general closed surfaces. For multiply-connected surfaces, we can directly decompose the surfaces along their critical trajectories. For general closed surfaces, the holomorphic quadratic differentials whose horizontal trajectories are closed loops induce the surface decomposition. The rationale of these properties are described as follows.

%In the CPP problem, obstacles in a complex surface are the spaces that cannot be passed. Therefore, the surface can be regarded as a multiply connected surfaces on which the obstacles are represented by holes. The path planning technique proposed in this work is applicable to general surfaces. We introduce the rationale for the closed surfaces with genus and multiply-connected surfaces.

\begin{definition}(Multiply-Connected Surface).
	Suppose $M$ is a surface of genus zero with multiple boundaries. Then $M$ is called a multiply-connected surface.
\end{definition}

%\subsubsection{Strebel Differentials} % (fold)
\medskip\noindent\textbf{Strebel Differentials.}
\label{ssub:strebel_differentials}
For a closed surface with genus $g>1$, holomorphic quadratic differentials induce the decomposition for the surface under some conditions. Those holomorphic quadratic differentials are called \emph{Strebel differentials}.
% Here we first introduce a class of holomorphic quadratic differentials, \emph{Strebel Differential}, which induces a natural decomposition of a closed surface with genus $g>1$. Then we discuss our decomposition method for any multiply-connected surface.

\begin{definition}(Strebel Differential\cite{strebel1984quadratic,douady1975density}).
	Suppose $\Phi_{s}$ is a holomorphic quadratic differential on a surface $R$ with genus $g>1$. $\Phi_{s}$ is called a \emph{Strebel differential}, if all of its regular horizontal trajectories are closed loops.
	% Given a holomorphic quadratic differential $\Phi_{s}$ on a Riemann surface $R$, if all of its horizontal trajectories are finite closed loops, then $\Phi_{s}$ is called a \emph{Strebel differential}.
\end{definition}

\begin{figure}[t]
%	\vspace*{4pt}
	\centerline{\includegraphics[width=0.5\columnwidth]{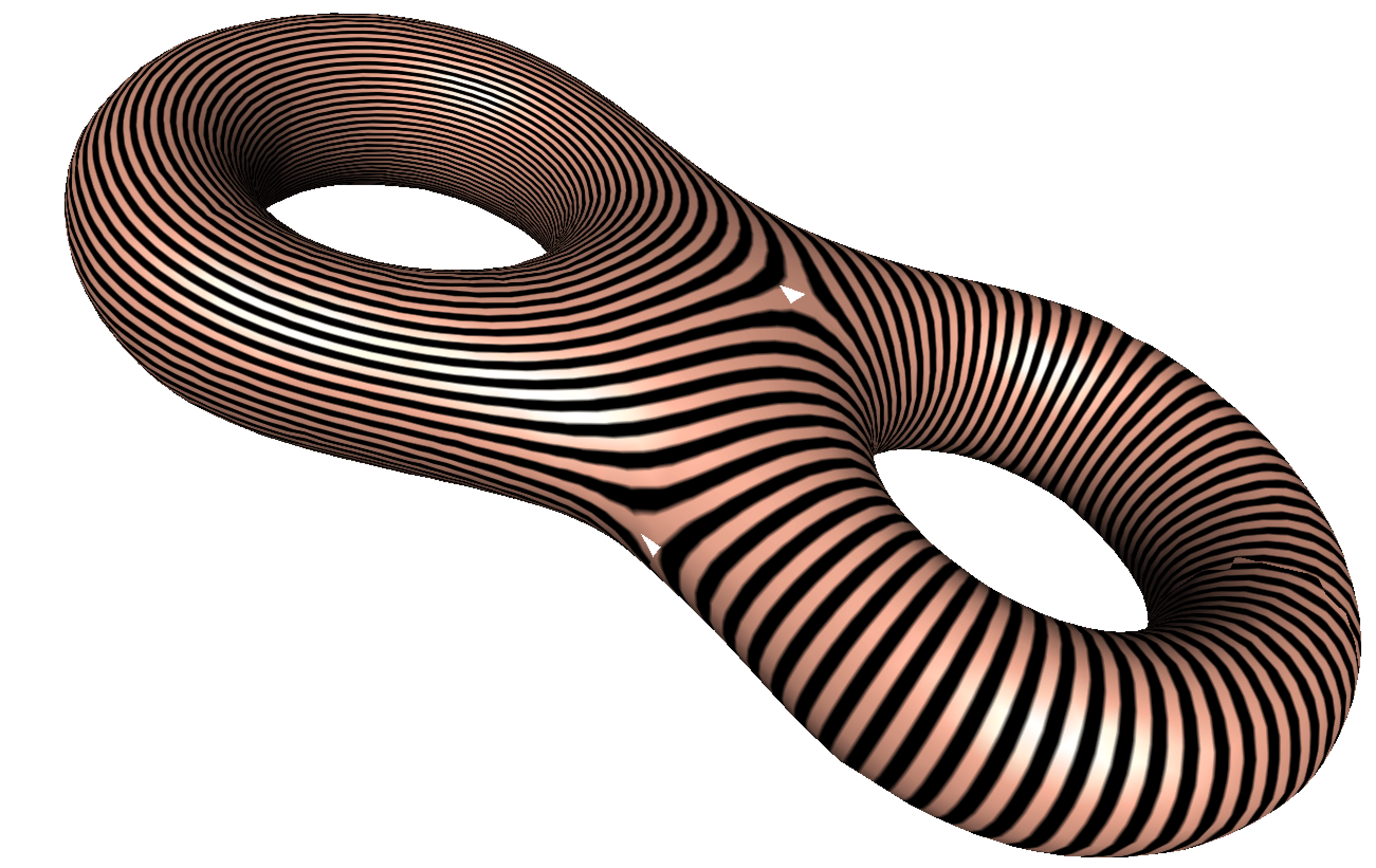}\includegraphics[width=0.5\columnwidth]{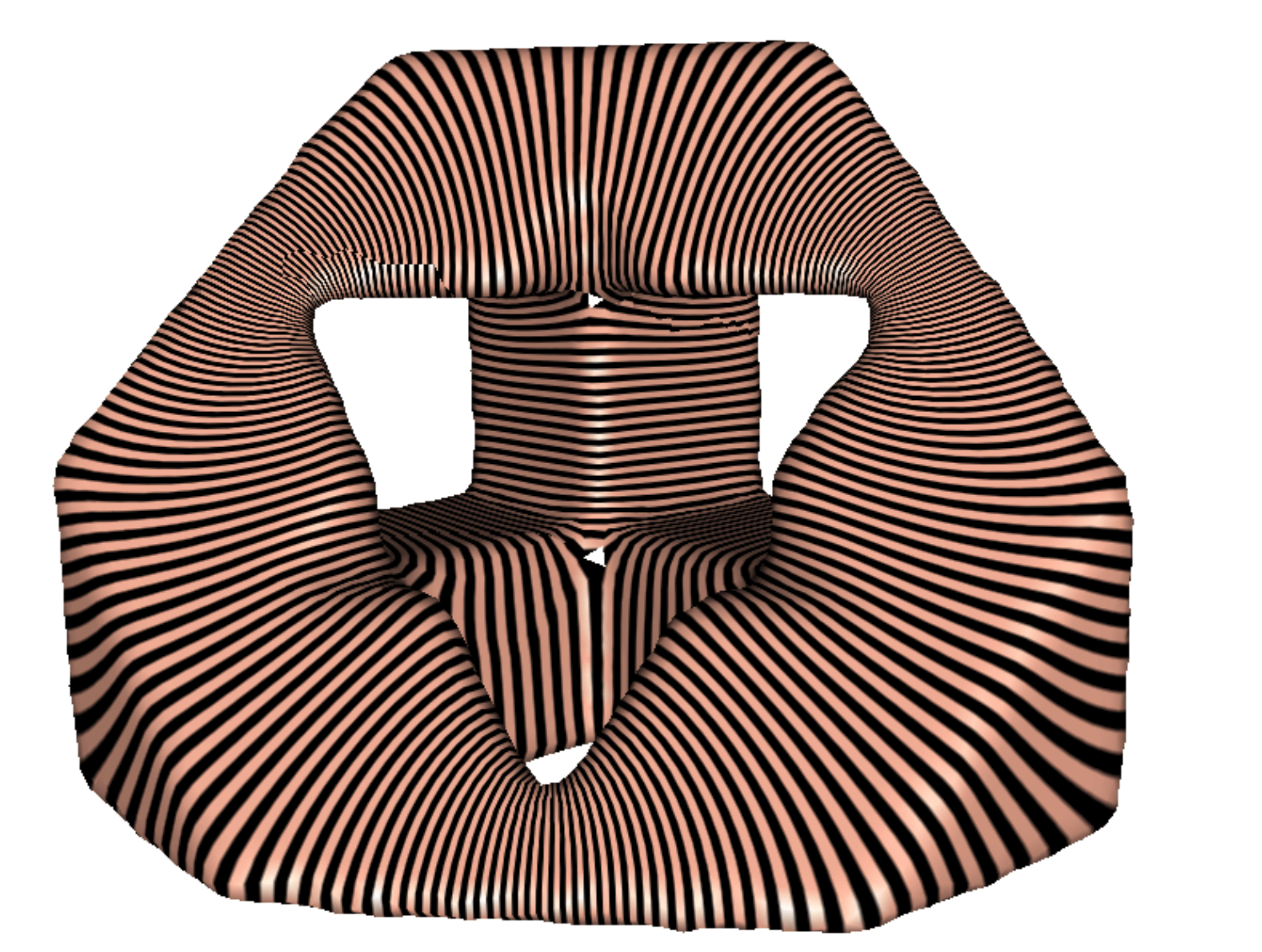}}
	\caption{The regular horizontal trajectories of a Strebel differential are closed loops on the surface.}\label{fig:strebel}
	\vspace*{-6pt}
\end{figure}

Notice that for a Strebel differential $\Phi_{S}$ on a closed surface $R$ with genus $g>1$, all the regular horizontal trajectories are closed loops as shown in Fig.~\ref{fig:strebel}. The set of critical trajectories together with the critical points form the critical graph $\Gamma$ of Strebel differential $\Phi_{s}$. The critical graph $\Gamma$ decomposes the surface $R$ into $3g-3$ topological cylinders~\cite{strebel1984quadratic}.

%\subsubsection{Symmetric Quadratic Differentials} % (fold)
\medskip\noindent\textbf{Symmetric Quadratic Differentials.}
\label{ssub:symmetric_quadratic_differentials}
For any given multiply-connected surface $M$ with $n>2$ boundaries, we can find a holomorphic quadratic differential which decomposes $M$ into $3n-3$ simply-connected surfaces $\{d_{1}, d_{2}, \dots, d_{3n-3}\}$. 

According to the symmetric image property~\cite{strebel1984quadratic}, $M$ and its double $\bar{M}$ form a symmetric surface $\tilde{M}=\{M\cup\bar{M}\}$ on which their corresponding boundaries are identified. Any holomorphic quadratic differential $\Phi$ on $M$ is reflected to $\bar{M}$. As a result, A symmetric surface $\tilde{M}$ is with a symmetric holomorphic quadratic differential $\tilde{\Phi}$. Because the boundaries $\partial{M}$ and $\partial{\bar{M}}$ are identified, each horizontal (vertical) trajectory $\gamma$ of $M$ and its symmetric trajectory $\bar{\gamma}$ of $\bar{M}$ are connected and form a closed loop.

The symmetric surface $\tilde{M}$ is, therefore, a closed surface with genus $g=n$. The holomorphic quadratic differential $\tilde{\Phi}$ on $\tilde{M}$ is a Strebel differential, which means the critical graph decomposes $\tilde{M}$ into $3n-3$ topological cylinders. Each cylinder $c_{i}$ is symmetric along the two curves which are some intervals of $\partial{M}$. That is to say, $c_{i}$ consists of two symmetric simply-connected domain $d_i$ and $\bar{d_i}$. By considering $\{d_{1}, d_{2},\dots, d_{3n-3}\}$, we can conclude that the holomorphic quadratic differential $\Phi$ decomposes $M$ into $3n-3$ simply-connected surfaces.

\section{Algorithm} % (fold)
\label{sec:algorithm}
% For a topological sphere, we can map it to a 2D plane by stereo-graphic projection.
The core idea of the proposed algorithm is the \emph{holomorphic quadratic differentials}, which induce surface parameterizations for general surfaces. In brief, holomorphic quadratic differentials inherently induce non-intersecting trajectories on a surface as shown in Figure~\ref{fig:strebel}. This property provides us enough freedom on manipulating the trajectories, and motivates us to develope our path planning algorithm.

Holomorphic quadratic differentials can be obtained by multiplying two holomorphic differentials. ~\ref{ssub:Holomorphic_diff} briefly lists the computational steps of holomorphic differentials. The parameterizations of holomorphic quadratic differentials should satisfy the property of being a curl-free vector field. It is challenging to control the numerical error around critical points due to the special local structure. As for our robot cover path, it avoids the zero points and all we need is to trace out the 3 critical trajectories through each critical point.

For a topological torus (closed surface with genus one) and an annulus, the holomorphic quadratic differentials and holomorphic differentials are equivalent. Therefore, by connecting each path induced by the trajectories of a holomorphic differential, a path planning is obtained. The algorithm described in this section focuses on the closed surfaces with genus $g>1$, and the multiply-connected surface with $n>2$ boundaries. For a closed surface with boundaries, we can double cover the surface to become a closed surface with genus $g>1$. Then the algorithm can be directly applied. 

\subsection{Discrete Approximation} % (fold)
\label{sub:discrete_approximation}
The mathematical concepts on smooth surfaces are now transformed to the numerical procedures on triangular meshes. A smooth surface is approximated by a piecewise linear triangle mesh $T$. The half-edge data structure is adopted in our implementation. We denote a vertex by $v_i$, a half-edge by $[v_{i},v_{j}]$, and an oriented triangle face by $[v_{i}, v_{j}, v_{k}]$.

A discrete differential is a function defined on the edge $\omega : E\rightarrow\mathbb{C}$. The integration of a discrete differential, $f:V\rightarrow\mathbb{C}$, gives a complex number or a $uv$-coordinate to each vertex.

\subsection{Algorithm Overview} % (fold)
\label{sub:algorithm_overview}
The following pipeline shows a summary of the main procedures of the path planning in this paper. The input is a triangular mesh of a closed surface with genus $g>1$, or a multiply-connected surface with $n>2$ boundaries. We first compute the \emph{holomorphic differential} basis on a surface, which is then used to compute the \emph{holomorphic quadratic differentials}. The holomorphic quadratic differential induces a global parameterization, and the resulting critical trajectories naturally decompose the surface into $3g-3$ ($3n-3$) sub-surfaces. For each sub-surface, we can compute a number of paths by tracing regular trajectories. The paths are concatenated together to become a zig-zag path on the sub-surface. Finally, we combine the sub-surfaces back to get a continuous path on the whole surface.

\begin{algorithm}
	\label{algo:quad_mesh}
    \SetKwInOut{Input}{Input}
    \SetKwInOut{Output}{Output}

    \Input{A triangle mesh $T$}
    \Output{A coverage path planning $P$ of $T$}
	% Glue two copies of $T$ along their corresponding boundaries. A symmetric closed surface $M$ with genus $g\geq 2$ is generated. The outer boundaries of the two copies of $T$ form a closed loop $C$ on $M$;
	
	Compute a holomorphic differential basis for $T$;
	
	Compute a holomorphic quadratic differential $\Phi$ for $T$;
	
	% Slice $M$ along $C$ and choose one of the multiple-connected annulus $\bar{T}$. The holomorphic quadratic differential $\bar{\Phi}$ on $\bar{T}$ is used to do path covering;
	
	Locate zero points of $\Phi$ on $T$;
	
	Trace the critical graph $\Gamma$ from zero points; 
	
	$T$ is decomposed along the critical graph $\Gamma$ and the sub-surfaces $T\backslash\Gamma=\{d_{1},d_{2},\cdots,d_{3n-3}\}$ are obtained. For each $\{d_{i}\}$, generate a path planning $P_{i}$;
	
	The path planning of the whole surface is formed by $P_{1}\cup P_{2}\cup \cdots\cup P_{3n-3}\cup\Gamma$
    \caption{Coverage path planning}
\end{algorithm}
% subsection algorithm_overview (end)

\subsection{Holomorphic Differentials} % (fold)
\label{ssub:Holomorphic_diff}
The computation of holomorphic differentials is to solve an elliptic partial differential equation on a triangle mesh using finite element method. The key step is to use piecewise linear functions defined on edges to approximate differentials. Furthermore, the differentials minimize the harmonic energy, the existence and the uniqueness are guaranteed by the Hodge theory~\cite{schoen1997lectures}. The following algorithm focuses on the closed surfaces with genus $g>1$. For a multiply-connected surface with $n>2$ boundaries, the algorithm is simplified to skip the computation of homology basis~\cite{Yin:2008gd}. Readers can refer to the works by Gu et al.~\cite{gu2002computing,gu2003global} for more details.

\begin{algorithm}
	\label{algo:holoform}
    \SetKwInOut{Input}{Input}
    \SetKwInOut{Output}{Output}

    % \underline{function Euclid} $(a,b)$\;
    \Input{A closed mesh $T$ with genus $g>1$}
    \Output{A holomorphic differential basis of $T$}
	%$\{\omega_{1}+\sqrt{-1}*\omega_{1}, \omega_{2}+\sqrt{-1}*\omega_{2},\cdots, \omega_{2g}+\sqrt{-1}*\omega_{2g}\}$
	Compute the homology group basis $\{\gamma_{1}, \gamma_{2},\cdots, \gamma_{2g}\}$ of $T$;
	
	Compute the dual cohomology group basis $\{\psi_{1}, \psi_{2},\cdots, \psi_{2g}\}$ of $T$;
	
	% , such that $\oint_{\gamma_i}\omega_{j}=\delta_{i}^{j}$, where
% 					\[
% 		\delta_{i}^{j} = \left\{\begin{array}{ll}
% 		1, & \mbox{if $i=j$} \\
% 		0, & \mbox{if $i\neq j$}.
% 		\end{array} \right.
% 		\]
		
	Compute the harmonic differential basis from the dual cohomology group basis $\{\psi_{1}, \psi_{2},\cdots, \psi_{2g}\}$ using heat flow method;
	%$\{\omega_{1}, \omega_{2},\cdots, \omega_{2g}\}$
	
	% For each harmonic differential base $\omega_{x}$, locally rotate by a right angle about the normal to obtain $\omega_{y}$. The pair $(\omega_{x},\omega_{y})$ forms a holomorphic differential
	For each harmonic differential base $\omega_{i}$, locally rotate by a right angle about the normal to obtain $\sqrt{-1}*\omega_{i}$. $\omega_{i}+\sqrt{-1}*\omega_{i}$ forms a holomorphic differential $\zeta_i$
	% \emph{Closedness: }$d\omega=0 \Rightarrow\omega(\partial[u,v,w])=\omega[u,v]+\omega[v,w]+\omega[w,u]=0$;
%
% 	\emph{Harmonity: }$\Delta\omega=0\Rightarrow$ $\sum_{[u,v]\in M}k_{u,v}\omega[u,v]=0$, where $k_{u,v}=-\frac{1}{2}(\cot\alpha+\cot\beta)$;
%
% 	Harmonic 1-form basis of $M$: $\{\omega_{1}, \omega_{2},\cdots, \omega_{2g}\}$;
%
% 	\emph{Conjugacy: }$*\omega_{i}=n\times\omega_{i}$

    \caption{Holomorphic differentials}
\end{algorithm}

% subsubsection subsubsection_name (end)

In the algorithm below, the holomorphic differential $\omega_{i}+\sqrt{-1}*\omega_{i}$ is denoted by $\zeta_{i}$, where $i\in \{1,2,\cdots,2g\}$.

\subsection{Holomorphic Quadratic Differentials} % (fold)
\label{sub:Holomorphic_quadratic_diff}
The holomorphic quadratic differentials on a surface can be obtained from the products of any two holomorphic differentials $\Phi=\{\zeta_{i}\cdot\zeta_{j}\}$, $i,j\in \{1,2,\cdots,2g\}$.
 
\begin{algorithm}
	\label{algo:hqd}
    \SetKwInOut{Input}{Input}
    \SetKwInOut{Output}{Output}

    \Input{A triangle mesh $T$ and two holomorphic differentials $\zeta_{i},\zeta_{j}$, $i,j\in \{1,2,\cdots,2g\}$}
    \Output{A holomorphic quadratic differentials $\Phi = \zeta_{i}\cdot\zeta_{j}$}
		% Compute the holomorphic differentials on $M$ by Algorithm ~\ref{algo:holoform}.
		
		Compute the products of the holomorphic differentials $\zeta_{i}\cdot\zeta_{j}$

    \caption{Holomorphic quadratic differential}
\end{algorithm}

\begin{figure}[htbp]
	\centering
% 	\vspace{-4mm}
	\includegraphics[width=1\columnwidth]{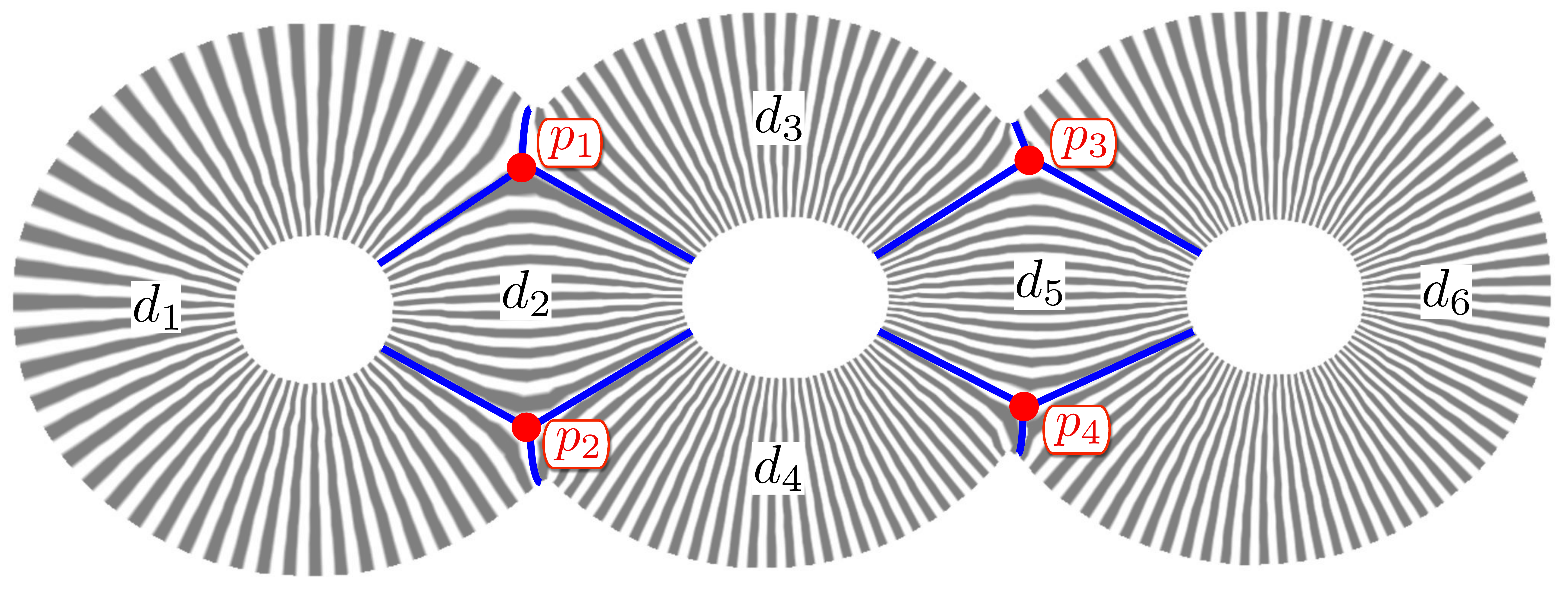}
% 	\vspace{-6mm}
	\caption{A three-hole donut with zero points($p_1 \sim p_4$) and simply-connected surfaces($d_1 \sim d_6$) decomposed by the critical trajectories(in blue).}
% 	\vspace{-4mm}
	\label{fig:tripledonut}
\end{figure}
\subsection{Surface Decomposition} % (fold)
For a closed surface with genus $g>1$, the surface decomposition is induced by Strebel differentials. Since holomorphic quadratic differentials $\zeta_{i}\cdot\zeta_{j}$ form a vector space, and Strebel differentials are the holomorphic quadratic differentials with closed horizontal trajectories. Therefore, a Strebel differential can be computed by the linear combination of holomorphic quadratic differentials. The surface is decomposed to $3g-3$ topological cylinders with two boundaries $\{c_{1}, c_{2}, \dots, c_{3g-3}\}$. For any multiply-connected surface with $n>2$ boundaries, the critical graph of a holomorphic quadratic differential decomposes the surface to $3n-3$ simply-connected surfaces $\{d_{1}, d_{2}, \dots, d_{3n-3}\}$.

In order to decompose the given surface along the critical graph of a computed holomorphic quadratic differential, we first locate the zero points on the surface. Then we trace the critical trajectories from the zero points. Figure~\ref{fig:tripledonut} illustrates the surface decomposition. For a surface with three holes (inner boundaries), there are four critical points (zero points) and six simply-connected domains.

\begin{algorithm}
	\label{algo:locatezeros}
    \SetKwInOut{Input}{Input}
    \SetKwInOut{Output}{Output}
    \Input{A holomorphic quadratic differential $\Phi$ on $T$}
    % \Output{A Strebel differential $\Phi_{s}$ on $M$ with colored vertices $\{p_{1},p_{2},\cdots,p_{4g-4}\}$}
	Given a vertex $v\in T$, find all vertices connecting to $v$ sorted counterclock-wisely, denoted as $w_{0},w_{1},\cdots,w_{n-1}$;
	
	Map $w_i$ to the plane using its natural coordinate $\xi(w_i):=\int^{w_i}_{v}\sqrt{\Phi}$;
	
	The points $\xi(w_0),\xi(w_1),\cdots,\xi(w_{n-1})$ form a planar polygon and the point $\xi(v)$ is inside this polygon. Compute the summation of the angles
	\begin{equation*}
		\sum_{i=0}^{n-1}\angle\xi(w_i)\xi(v)\xi(w_{i+1}),
	\end{equation*}
	where $w_{n}=w_{0}$. If the summation is $2\pi$, then $v$ is a regular point; if the summation is no less than $3\pi$, then $v$ is a zero point
    \caption{Locate zero points}
\end{algorithm}

\begin{algorithm}
	\label{algo:trace1}
    \SetKwInOut{Input}{Input}
    \SetKwInOut{Output}{Output}
    \Input{A holomorphic quadratic differential $\Phi$ on $T$ and a zero point $p_i$ of $\Phi$}
	% \Output{A mesh $M'$ refined from $M$ along the critical trajectories}
	For $p_i\in T$, find all faces adjacent to $p_i$ sorted counterclock-wisely, denoted as $f_{0},f_{1},\cdots,f_{n-1}$;
	
	For each vertex $w_j$ of $f_k$, map $w_j$ to the plane using its natural coordinate $\xi(w_j):=\int^{w_j}_{p_i}\sqrt{\Phi}$. The computation of natural coordinates is shown in Algorithm~\ref{algo:choosesign};
	
	The points $\xi(w_0), \xi(w_1), \xi(w_2)$ form a planar triangle, where $w_{0}=p_i$ and $w_0$ is mapped to the origin. Let $y_{1},y_{2}$ be the imaginary natural coordinates of $\xi(w_1), \xi(w_2)$ respectively. If $y_{1}y_{2} < 0$, then the planar triangle, denoted as $\Delta_{\xi}$, is passed by a critical trajectory $\gamma$;
	
	Compute the natural coordinates starting from $\Delta_{\xi}$. Find all of the parameterized triangles passed by $\gamma$. For a $g>1$ closed mesh, trace $\gamma$ until hitting a zero point; For a multiply-connected mesh with $n>2$ boundaries, trace $\gamma$ until hitting a boundary;
	
	Interpolate the critical trajectory $\gamma$, by which the planar triangles are passed
    \caption{Trace critical trajectories of $\Phi$}
\end{algorithm}

\begin{algorithm}
	\label{algo:choosesign}
    \SetKwInOut{Input}{Input}
    \SetKwInOut{Output}{Output}
    \Input{A holomorphic quadratic differential $\Phi$ on $T$}
	Given a face $f\in T$, compute $\sqrt{\Phi}$. For each edge $e$ of $f$, the sign of $\sqrt{\Phi}$ is decided to satisfy $\oint_{f}\sqrt{\Phi}=0$ because $\sqrt{\Phi}$ is a curl free vector field;
	% That is, $\sum_{i=1}^{3}duv=0$;
	
	For each vertex $v$ of $f$, compute the natural coordinate by the integration $\int\sqrt{\Phi}$
    \caption{Compute the natural coordinate of a holomorphic quadratic differential}
\end{algorithm}

\subsection{Coverage Path} % (fold)
The non-intersecting trajectories of holomorphic differential $\Phi$ give the paths for our coverage path planning. Here we take a multiply-connected surface $M$ as an example. Let the outer boundary of $M$ denoted by $l_1$, the inner boundaries denoted by $\{l_{2}, \cdots, l_{n}\}$, and the boundaries of the decomposed simply-connected surfaces denoted by $\{\partial d_{1}, \partial d_{2},\cdots, \partial d_{3n-3}\}$. Given any density step $\epsilon >0$, if $l_{1}\cap\partial d_{i}\neq\emptyset$, then we trace a regular trajectory for each density step $\epsilon$ along $l_{1}\cap\partial d_{i}$. Otherwise, there exists an inner boundary $l_i$ such that $l_{i}\cap\partial d_{i}\neq\emptyset$, and we trace a regular trajectory for each density step $\epsilon$ along $l_{i}\cap\partial d_{i}$. Once the paths are generated, we can simply connect the path together to form a zig-zag path.

% The critical trajectories decomposes a surface into a regular number of sub-surfaces. For any sub-surface obtained from the surface decomposition, the paths are generated by tracing a number of regular horizontal trajectories. 
\begin{figure}[htbp]
	\centering
% 	\vspace{-4mm}
	\includegraphics[width=0.45\columnwidth]{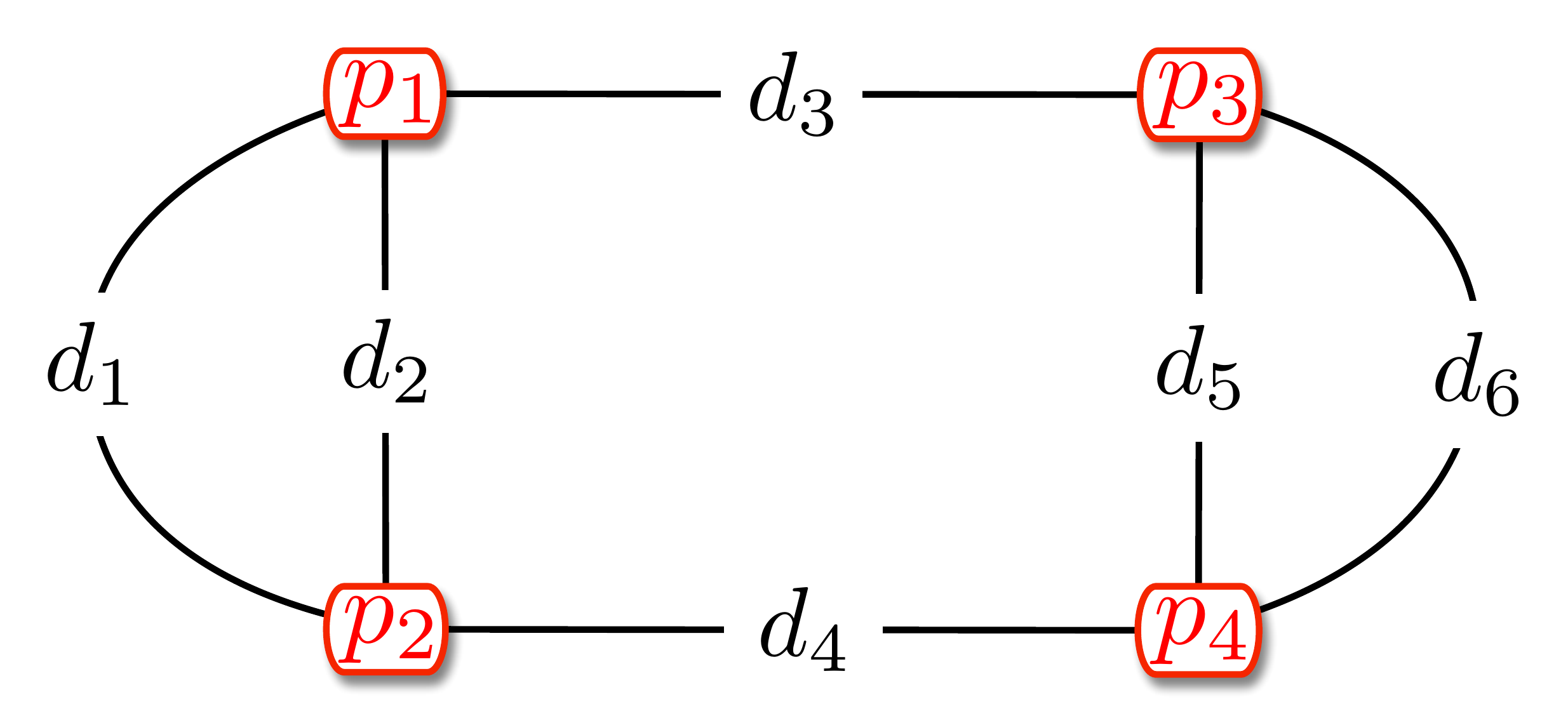}
% 	\vspace{-6mm}
	\caption{The dual graph of Figure~\ref{fig:tripledonut}. Here the zero points($p_1 \sim p_4$) are treated as nodes, and the decomposed simply-connected surfaces($d_1 \sim d_6$) that touched these points represent edges between nodes.}
% 	\vspace{-4mm}
	\label{fig:dual_graph}
\end{figure}

\subsubsection{Euler Cycle on Surface} % (fold)
\label{ssub:euler_cylce}
Based on our surface decomposition scheme, we discover that the zero points and the sub-surfaces can be converted to a dual graph $G_{M}$. That is, each zero point is dual to a node and each sub-surface is dual to an edge. Moreover, the necessity of visiting every sub-surface inspires the idea of finding an Euler cycle of $G_M$. By doubling each edge, it is guaranteed to find an Euler cycle which promises the visiting of every sub-surface.

\begin{figure}[htbp]
	\centering
%	\vspace{-4mm}
	\includegraphics[width=0.55\columnwidth]{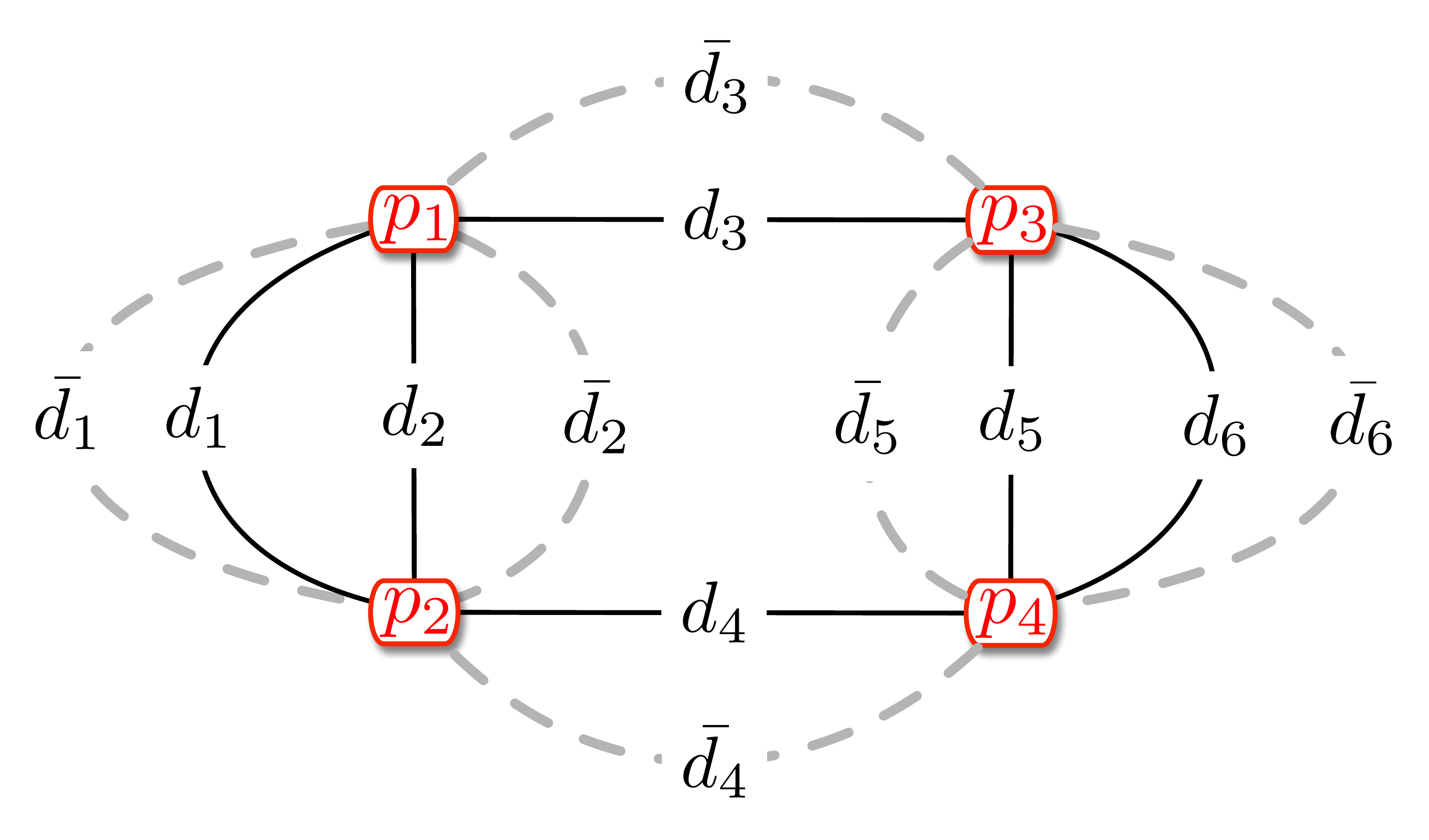}
% 	\vspace{-6mm}
	\caption{The doubled dual graph of Figure~\ref{fig:tripledonut}. We simply double each edge in the original dual graph(Figure~\ref{fig:dual_graph}).}
% 	\vspace{-4mm}
	\label{fig:dual_graph_doubled}
\end{figure}

\begin{figure}[htbp]
	\centering
%	\vspace{-4mm}
	\includegraphics[width=0.55\columnwidth]{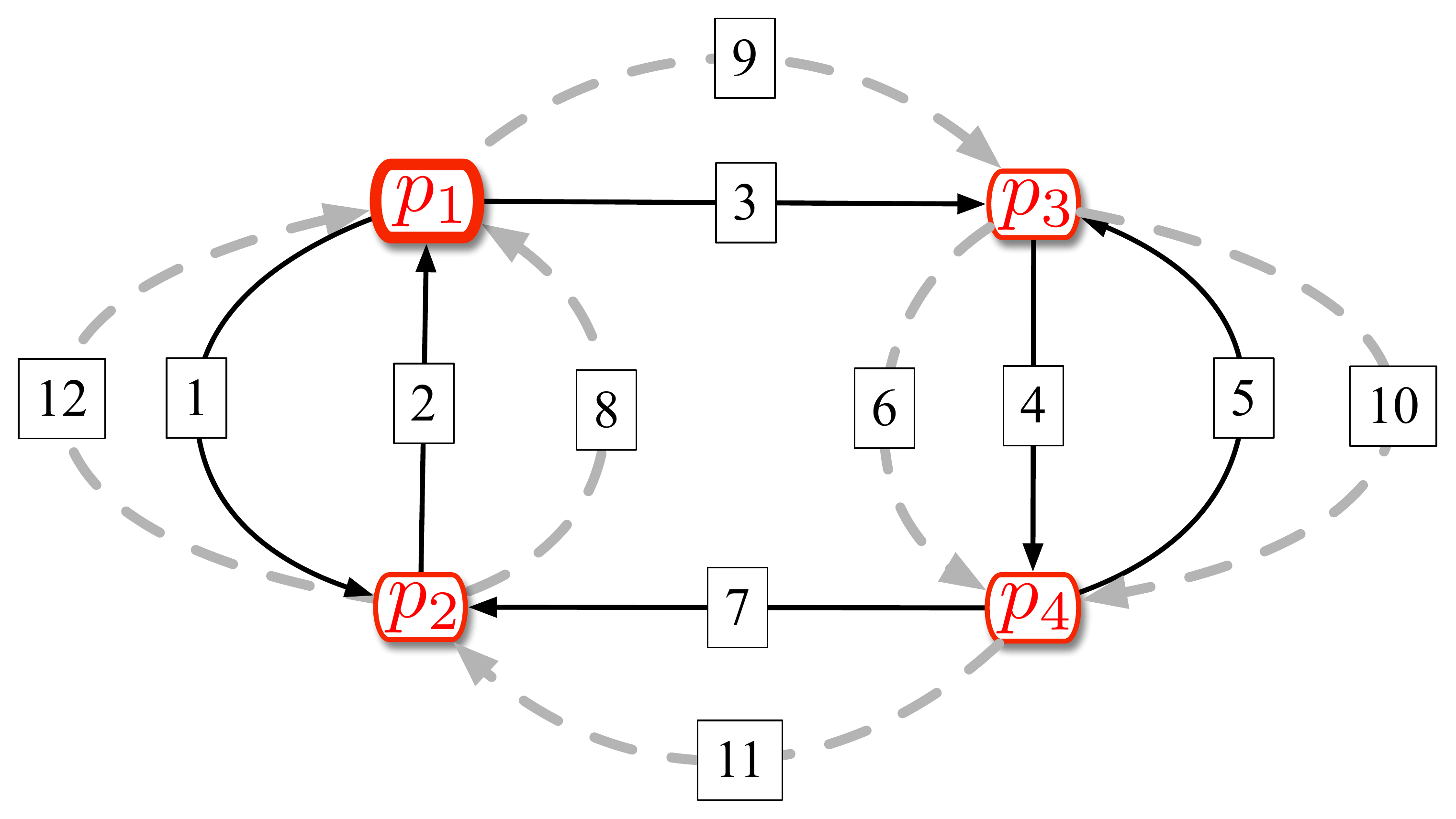}
% 	\vspace{-6mm}
	\caption{An Euler cycle example of the doubled dual graph of Figure~\ref{fig:tripledonut}. The cycle starts from $p_1$, and travels through the arrowed edge with number $1 \sim 12 $, and finally goes back to $p_1$. } 
% 	\vspace{-4mm}
	\label{fig:euler_cycle}
\end{figure}

Here we take the surface shown in Figure~\ref{fig:tripledonut} as an example. Its dual graph is illustrated in Figure~\ref{fig:dual_graph}. Each zero point $p_i$ is dual to a node, and each decomposed simply-connected surface $d_j$ is dual to an edge. Figure~\ref{fig:dual_graph_doubled} shows the doubled dual graph of Figure~\ref{fig:dual_graph}. For each edge $d_j$, the doubled edge $\bar{d}_j$ is created. Figure~\ref{fig:euler_cycle} shows an Euler cycle of the doubled dual graph of Figure~\ref{fig:tripledonut}. On the dual graph, Euler cycle makes the navigation start and end at the same point.

\subsubsection{Path Interlacement} % (fold)
\begin{figure}[htbp]
	\centering
% 	\vspace{-4mm}
	\includegraphics[width=0.3\columnwidth]{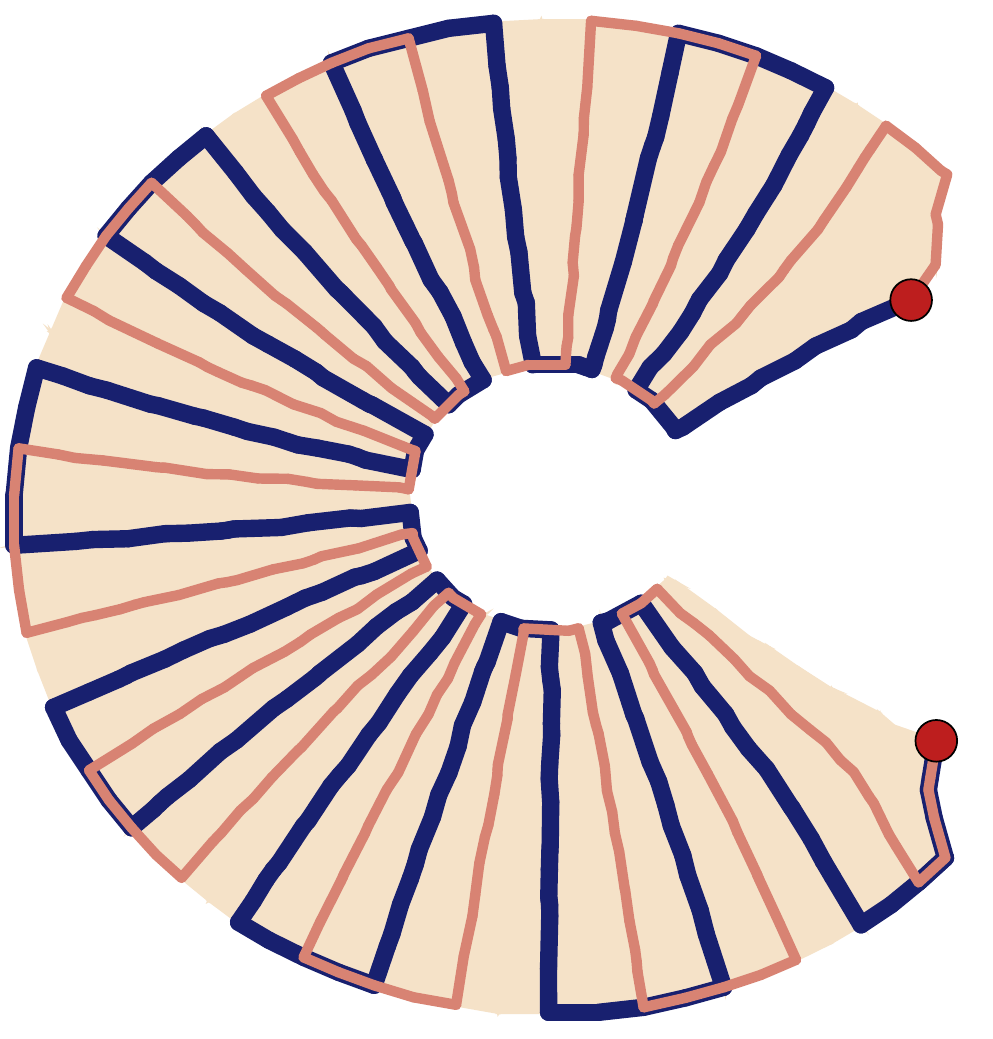}
% 	\vspace{-6mm}
	\caption{An Example of paths on the simply-connected domain $d_1$ in Figure~\ref{fig:tripledonut}. There are two paths traveling from one critical points to the other, labeled as blue and orange. When a robot travels between two zero points, it can choose a color on one way (as $d_1$ in Figure~\ref{fig:dual_graph_doubled}) and the other color on the other(as $\bar{d_1}$ in Figure~\ref{fig:dual_graph_doubled}), hence provide required path density for coverage.} 
% 	\vspace{-4mm}
	\label{fig:d1}
\end{figure}

\label{ssub:interlace}
Euler cycle of the dual graph of a surface implies that every sub-surface is visited twice. By interlacing two zig-zag paths with same density step, each sub-surface can be covered nicely. Figure~\ref{fig:d1} illustrates the interlacing paths on the simply-connected domain $d_1$ in Figure~\ref{fig:tripledonut}. There are two paths traveling from one zero points to the other, labeled as blue and orange. When a robot travels between two zero points, it can choose a color on one way (as $d_1$ in Figure~\ref{fig:dual_graph_doubled}) and the other color on the other(as $\bar{d_1}$ in Figure~\ref{fig:dual_graph_doubled}), hence provide required path density for coverage.

\begin{figure}[htbp]
	\centering
% 	\vspace{-4mm}
	\includegraphics[width=0.8\columnwidth]{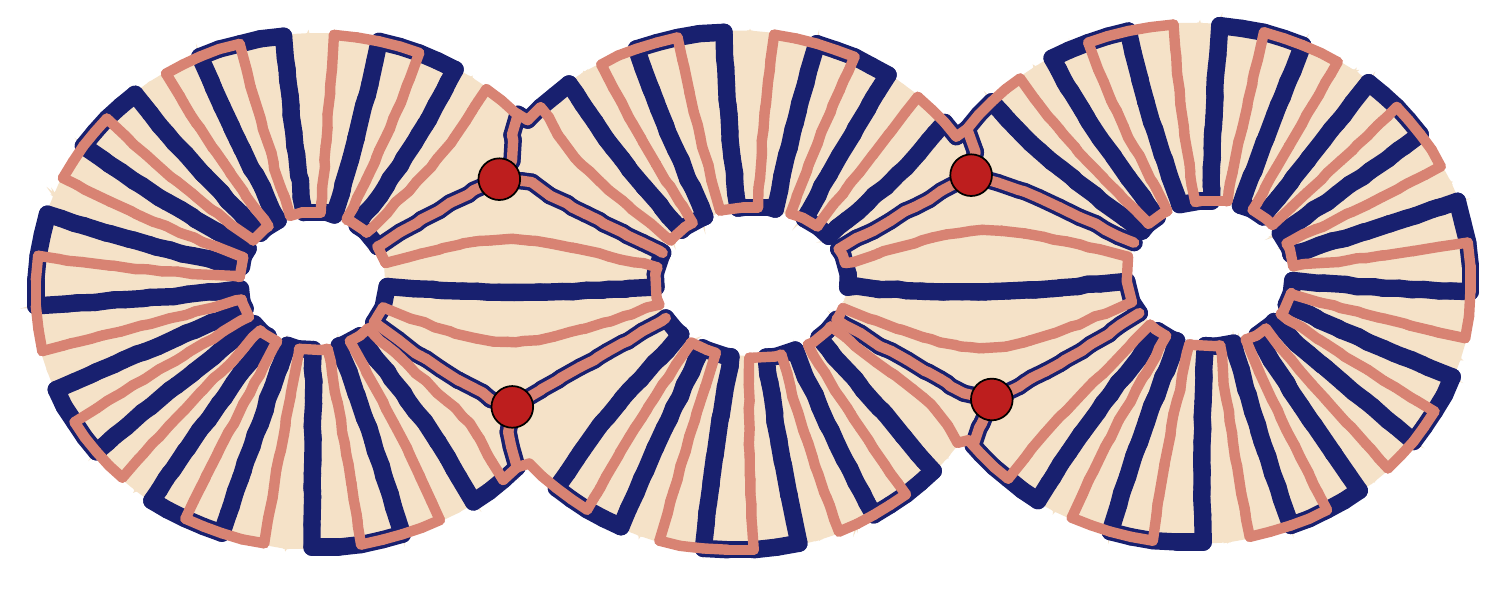}
% 	\vspace{-6mm}
	\caption{Coverage path for a three-hole donut.} 
% 	\vspace{-4mm}
	\label{fig:whole}
\end{figure}

Between the adjacent simply-connected domains, a robot can travel along the critical graph and transfer from one domain to another. By following the path interlacement and the Euler cycle scheme, the coverage path for the whole surface is performed. Figure~\ref{fig:whole} exhibits the coverage path result for a surface with three inner boundaries.

\begin{figure}[htbp]
	\centering
	%\vspace{-2mm}
	\subfigure[$\epsilon=8$, $\delta=0.005$ \hspace{40mm} $\epsilon=8$, $\delta=0.01$]{
		\label{fig:dount:subfig:8}
		\includegraphics[width=0.5\columnwidth]{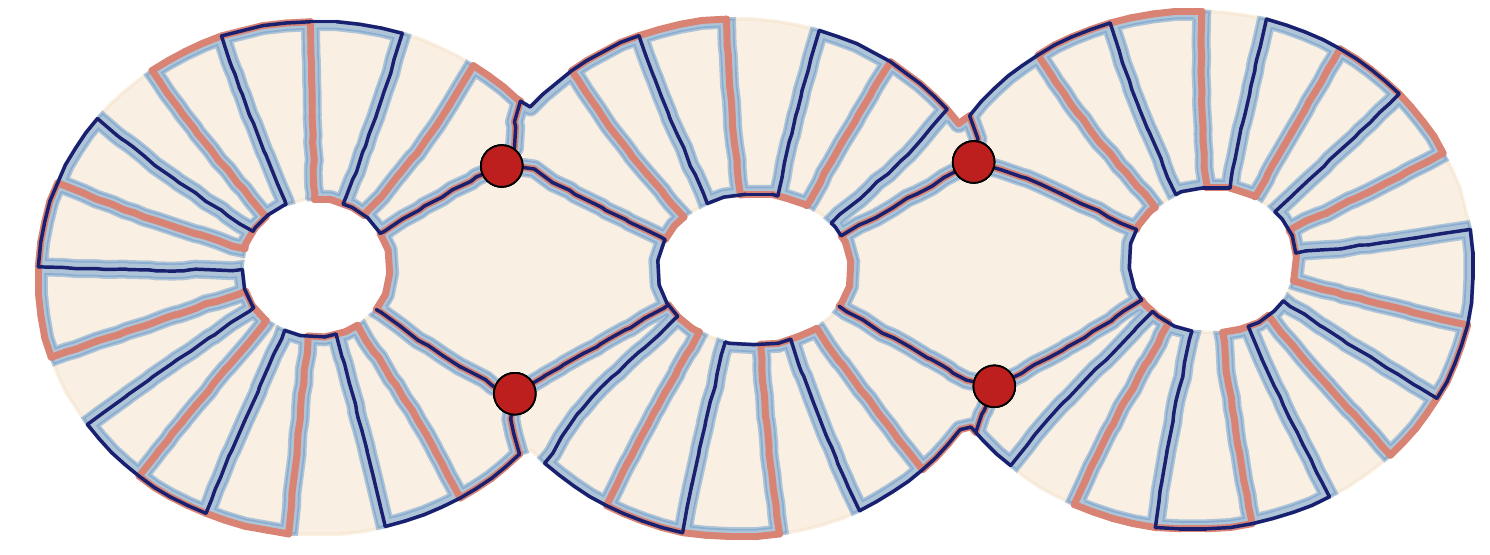}
		\includegraphics[width=0.5\columnwidth]{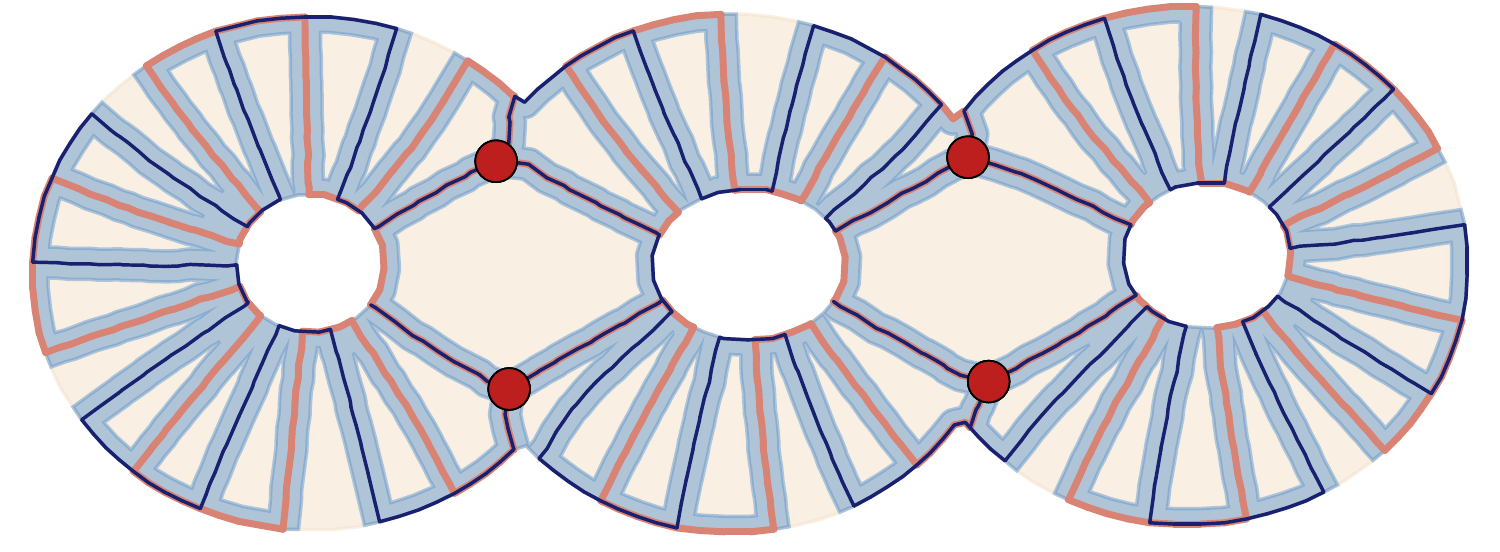}}
	\subfigure[$\epsilon=4$, $\delta=0.005$ \hspace{40mm} $\epsilon=4$, $\delta=0.01$]{
		\label{fig:dount:subfig:4}
		\includegraphics[width=0.5\columnwidth]{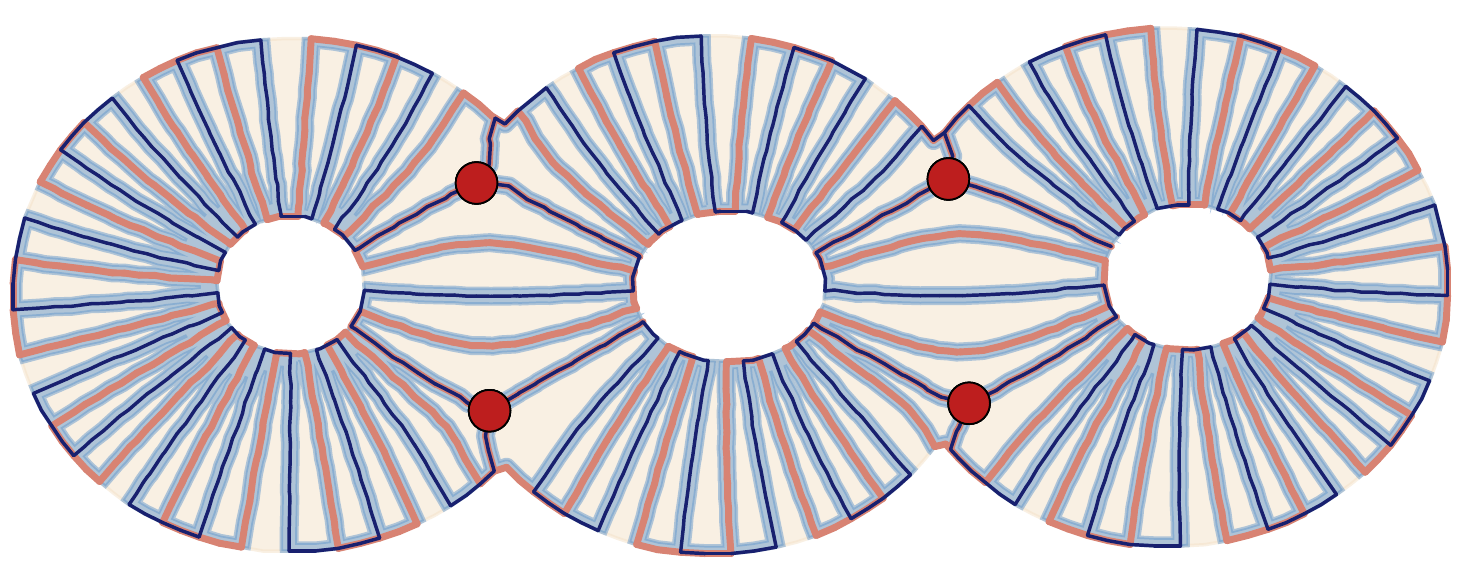}
		\includegraphics[width=0.5\columnwidth]{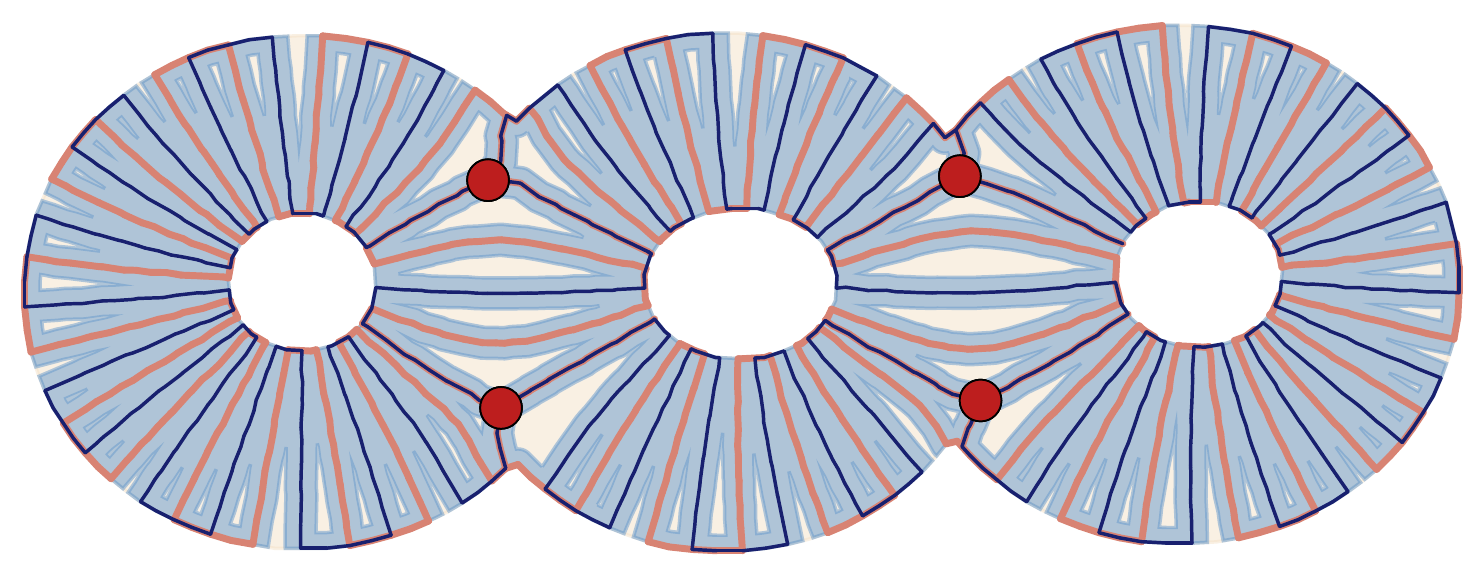}}
	\subfigure[$\epsilon=2$, $\delta=0.005$ \hspace{40mm} $\epsilon=2$, $\delta=0.01$]{
		\label{fig:dount:subfig:2}
		\includegraphics[width=0.5\columnwidth]{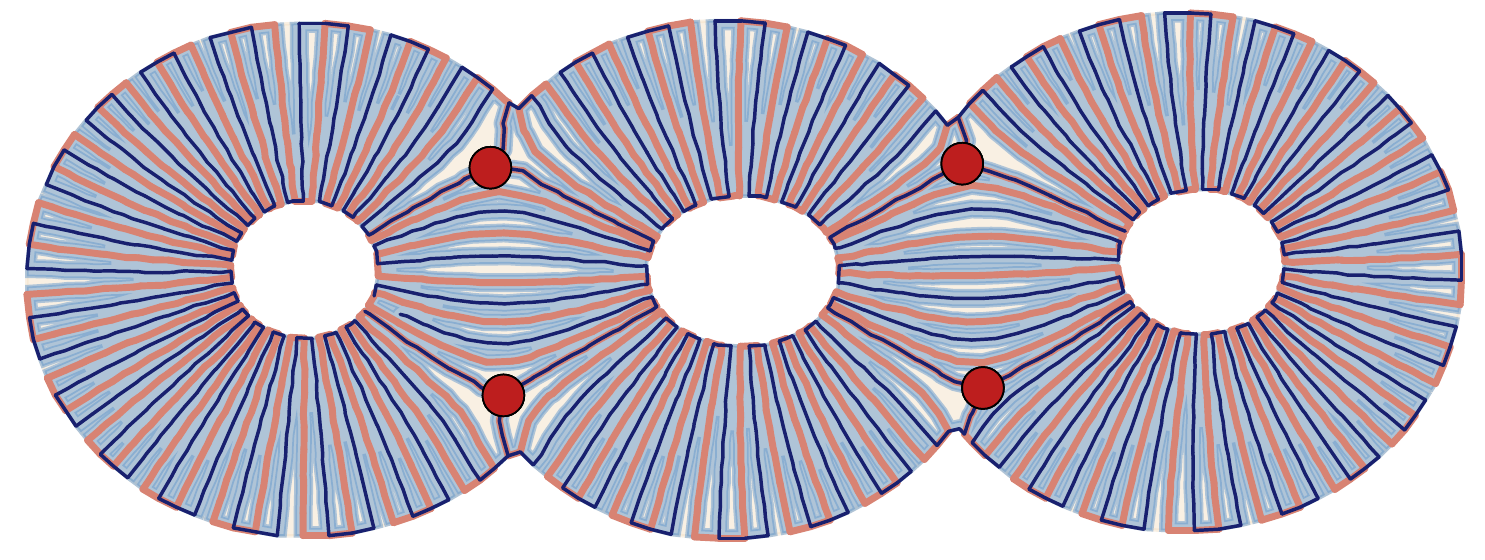}
		\includegraphics[width=0.5\columnwidth]{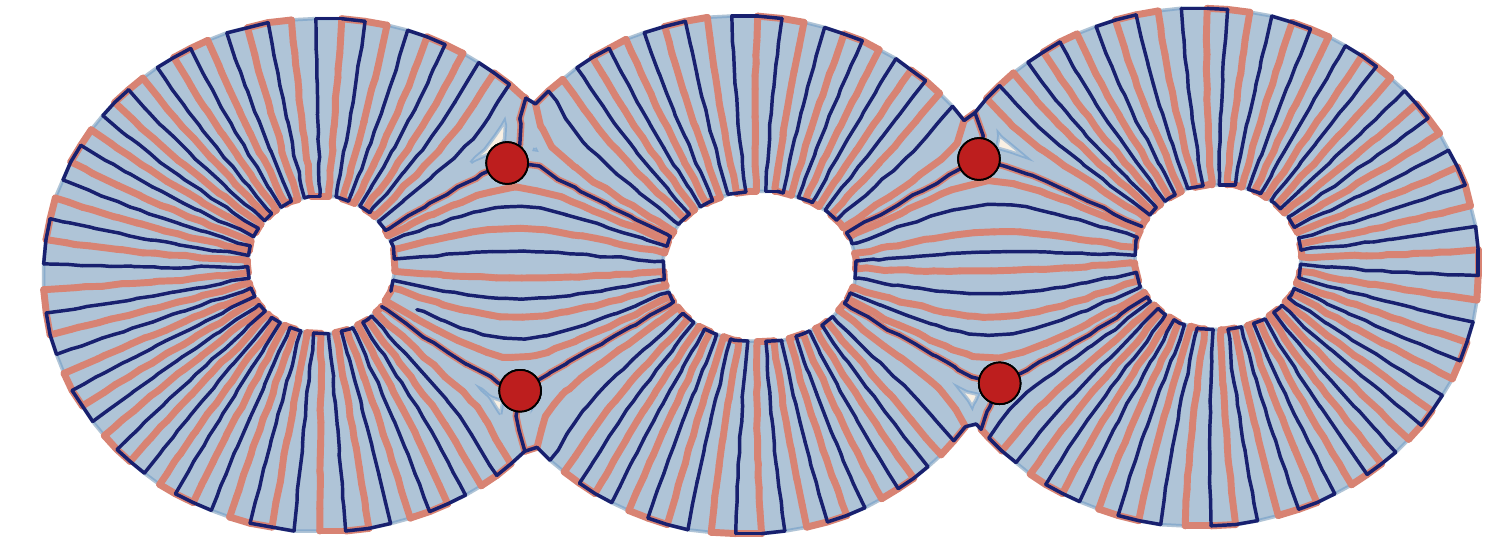}}
	% \vspace{-2mm}
	\caption{The path coverage of three holes donut with different density step $\epsilon$ and $\delta$. The orange and blue lines are the coverage paths, the singular points are labeled as red. Here the covered area is with light blue color. Notice that $\epsilon$ is inversely proportional to the path density.}
	% \vspace{-4mm}
	\label{fig:dount} %% label for entire figure
\end{figure}

\begin{figure}[htbp]
	\centering
% 	\vspace{-4mm}
	\includegraphics[width=0.8\columnwidth]{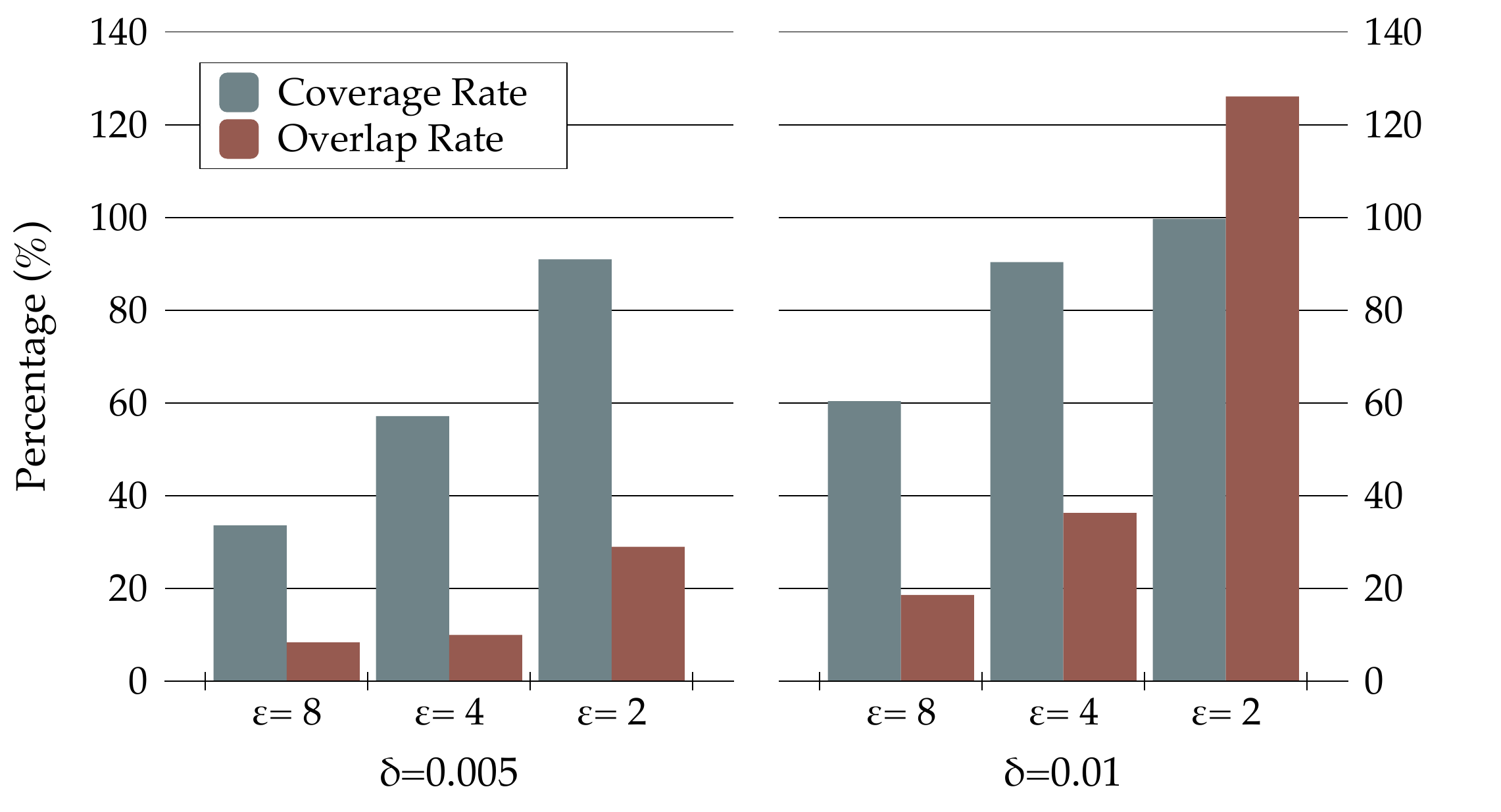}
% 	\vspace{-6mm}
	\caption{A comparison of density step $\epsilon$ and robot radius $\delta$ with coverage and overlap rate. Notice that $\epsilon$ is inversely proportional to the path density.} 
% 	\vspace{-4mm}
	\label{fig:coverage}
\end{figure}

\section{Experimental Results} % (fold)
\label{sec:exp}

\begin{figure}[htbp]
	\centering
	% \vspace{-6mm}
	\includegraphics[width=0.6\columnwidth]{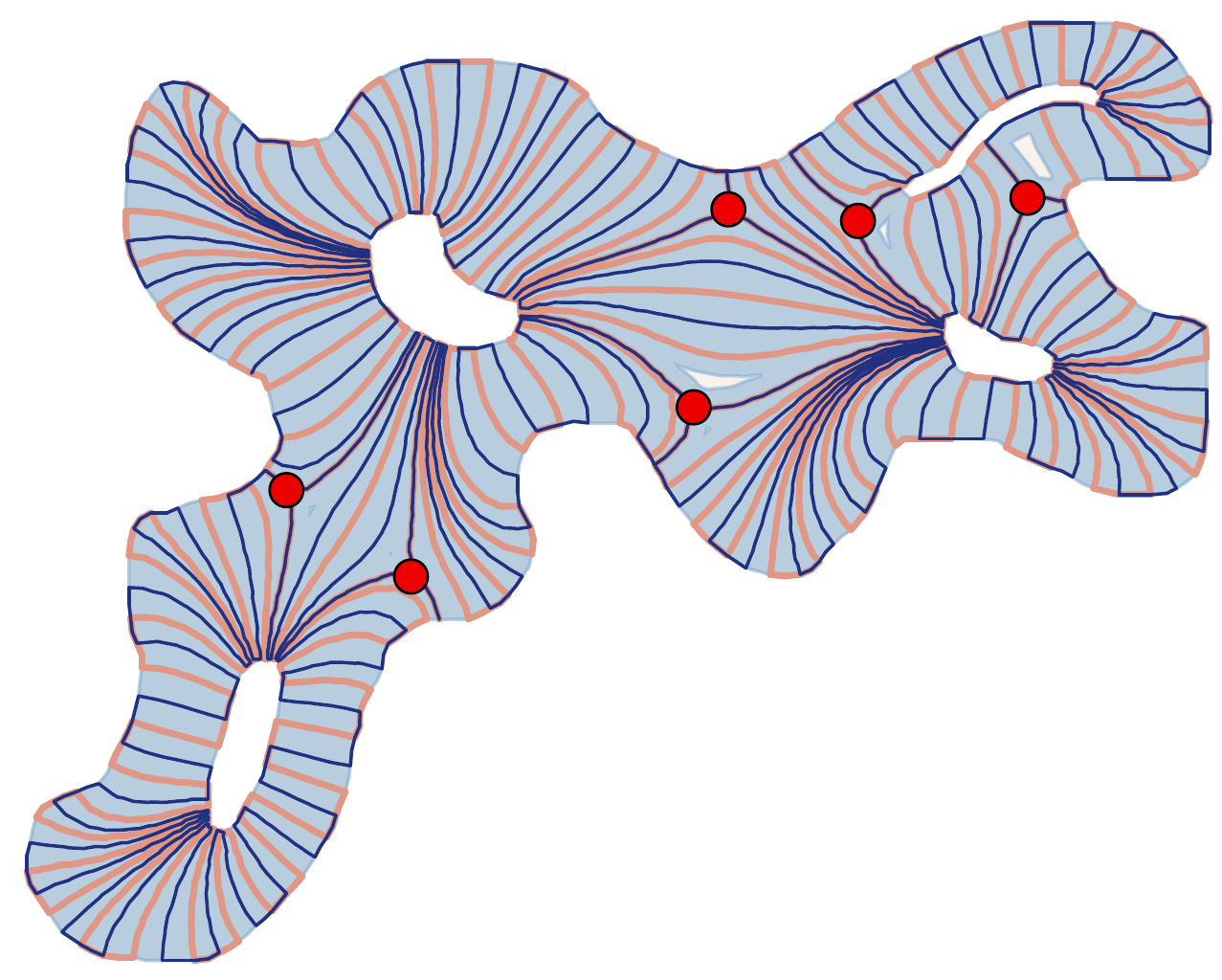}
% 	\vspace{-6mm}
	\caption{Example of a coverage path with a four-hole non-convex domain.} 
	% \vspace{-4mm}
	\label{fig:monster}
\end{figure}

\begin{figure}[htbp]
	\centering
	% \vspace{-6mm}
	\subfigure[]{
		\label{fig:terrain:subfig:1}
		\includegraphics[width=0.45\columnwidth]{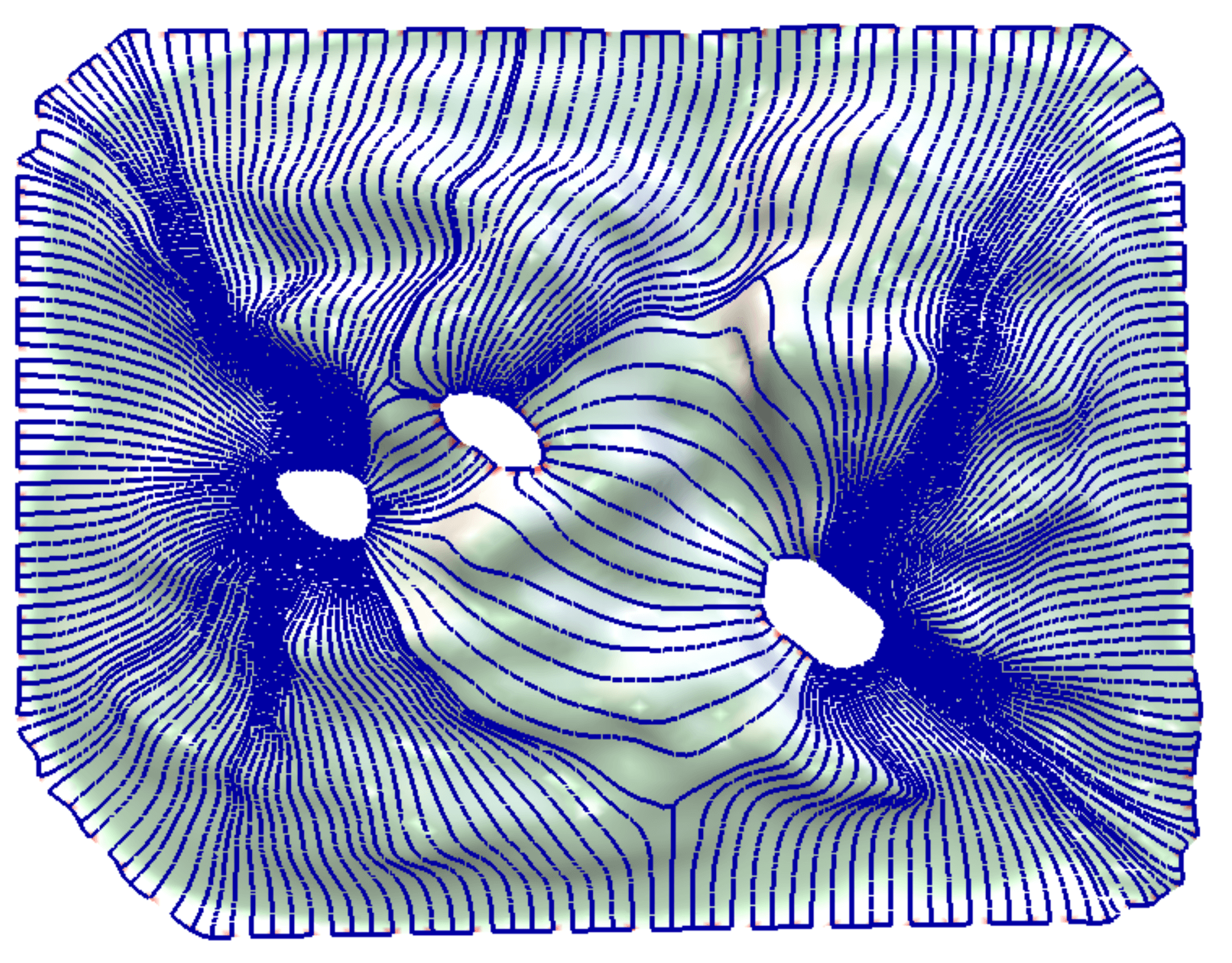}}
	% \hspace{0.0in}
	\subfigure[]{
		\label{fig:terrain:subfig:2}
		\includegraphics[width=0.45\columnwidth]{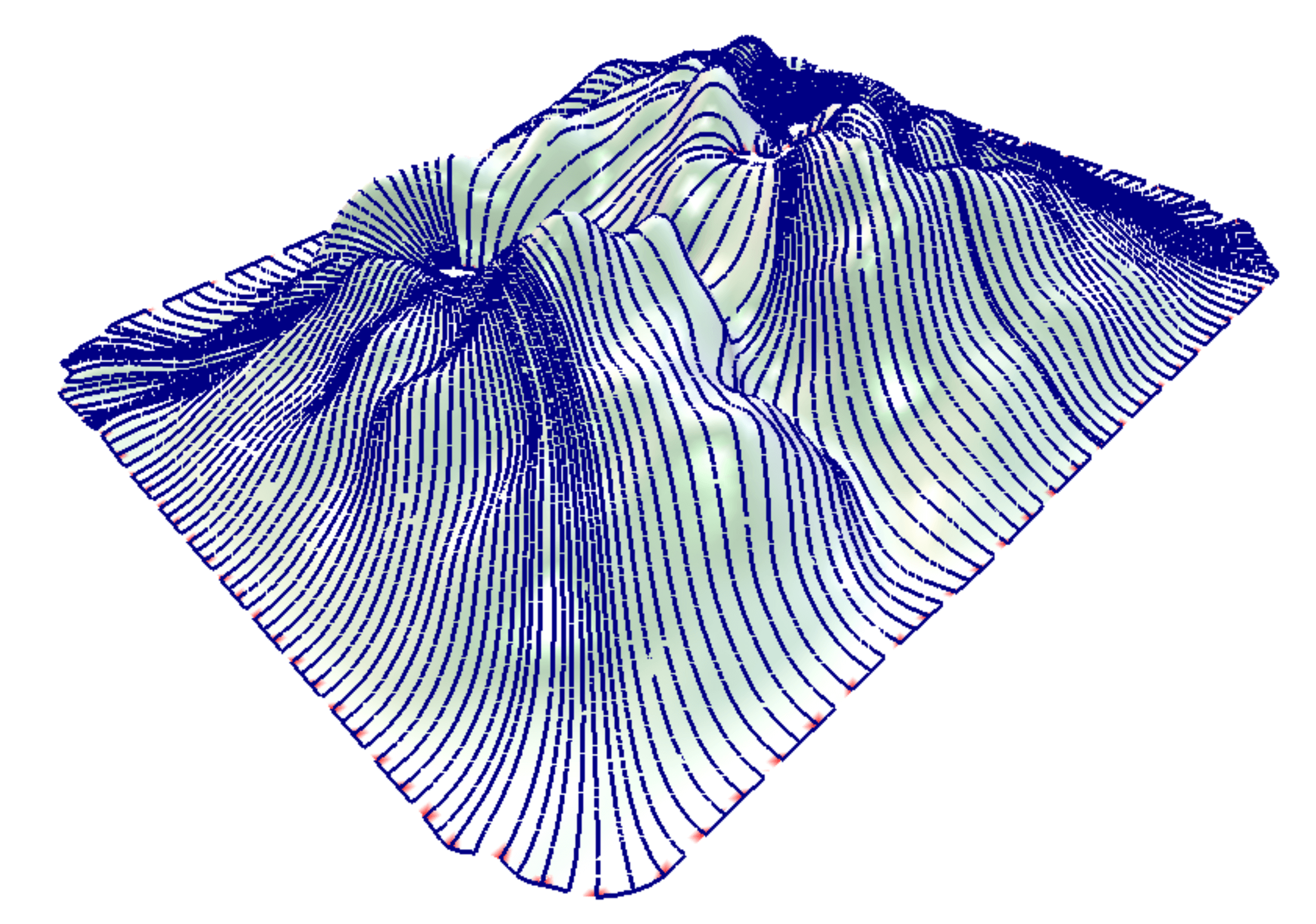}}
	% \vspace{-2mm}
	\caption{Example of a 3-D terrain with three lakes. The lakes are represented by empty holes.}
	% \vspace{-4mm}
	\label{fig:terrain} %% label for entire figure
		
\end{figure}

We evaluate our algorithm on various surfaces, and analyze the influence of different density step on coverage.
We first demonstrate our algorithm on a 2D three holes donut as in Figure~\ref{fig:tripledonut}. The coverage path result is displayed in Figure~\ref{fig:dount}. Here the robot covered area is colored with light blue. We fix the robot radius as $\delta$, and the density step $\epsilon$ represents the step distance on the outer boundary. Therefore, the bigger $\epsilon$ brings the sparser coverage paths. Notice that even smaller $\epsilon$ brings better coverage, but it also comes with a price of overlapped coverage. A comparison of the tradeoff between $\epsilon$, $\delta$, coverage rate and overlap rate is as Figure~\ref{fig:coverage}. As expected, the result shows that the overlap is more obvious on larger robot radius $\delta$ and denser density step $\epsilon$. Rather than the standard 2D domain, our algorithm is also suitable for complex 2D domain and 3D terrain with holes, the result of covering path is demonstrated in Figure~\ref{fig:monster} and Figure~\ref{fig:terrain}.

% section evaluation (end)

\section{Related Work}
 
% \noindent\textbf{Path Cover Problem.}
This problem has been studied extensively and one can refer to nice surveys~\cite{Galceran:2013kg,Choset:2001vf} for past work in this area. 
%One of the most important challenges in this area is to cover a generalized domain, for example, a 2D domain with non-convex boundary and obstacles, or a 3D surface domain with holes. 
For 2D domains, most works use a cell decomposition to decompose the domain into simple shapes. Popular cell decomposition includes classical trapezoid decomposition\cite{DeCarvalho:1997bs,Oksanen:2009gz} and boustrophedon cellular decomposition~\cite{Choset:1998cha,Garcia:2004kd,Xu:2011cl}, Morse decomposition~\cite{Acar:2002fx,Galceran:2012ix}, slice decomposition~\cite{Wong:2003fk}, various grid-based algorithms~\cite{Kapanoglu:2012bq,Zelinsky:1993te,Cai:2014un,Bhattacharya:2013bj}, etc. Most of these approaches are concerned of producing a small number of cells in the decomposition, and whether the decomposition can be done in the online setting (when the target domain is unknown and to be discovered). 
Other methods include applying spanning tree coverage\cite{Gabriely:2001gb,Zheng:2005kh} and neural network based coverage\cite{Luo:2002fa,Yang:2004gq,Yan:2012dk}. %\textbf{Add a few more sentences for the above two methods?}

Coverage path planning for surfaces in 3D is less investigated. Hert \etal~\cite{Hert:1996fn} considered coverage of a projectively planar 3D volume, they project the domain in 2D and then take advantages of the 2D planar terrain-covering algorithm to solve the problem.
Atkar \etal~\cite{Atkar:2001wz} extended the Morse decomposition to non-planar surfaces but did not consider obstacles. 
In \cite{Bhattacharya:2014vs} Bhattacharya et al. extended their grid-based algorithm\cite{Bhattacharya:2013bj} into 3D cased; they first separated the domain into voronoi cells, then handled them by multiple robots. In \cite{Jin:2011cl,Galceran:2013ds}, the authors proposed a lawnmower type of algorithm on 3D planar domain, but the results only show terrains with boundary and without obstacles. More heuristic algorithms are adopted in application scenarios as~\cite{Cheng:2008fn,Galceran:2013ds,Jin:2011cl}.

The one most relevant was our earlier work for generating a space filling curve~\cite{ban13topology}. However, the focus in~\cite{ban13topology} was to find a curve with progressive density -- that is, we want a path such that the distance from any point to the path to be shrinking progressively when the path gets longer. The same as in a followup work~\cite{li15space}. Although quadratic differentials were also used in~\cite{li15space} but both the theory and the algorithms for generating the curves are totally different from here.

The coverage path problem is also related to various traveling salesman problem (TSP with neighborhoods~\cite{Arkin:2000ir}), the lawnmower problem (full cover of a region by a path with minimum length)~\cite{Arkin:1994ds}, and the sweeping path problem (full coverage by a robot arm of fixed geometric degree of freedom)~\cite{Kim:2003gx}. Since these problems are sufficiently different we skip the results here. 

%TSPN: E.M. Arkin, R. Hassin, Approximation algorithms for the geometric covering salesman problem, Discrete Applied Mathematics 55 (3) (1994) 197–218.
%LawnMower: E.M. Arkin, S.P. Fekete, J.S. Mitchell, Approximation algorithms for lawn mowing and milling, Computational Geometry 17 (1–2) (2000) 25–50. URL: http://www.sciencedirect.com/science/article/pii/S0925772100000158.

%For 3D domains, a related problem is called sweeping path problem\cite{Kim:2003gx}. Given a general surface, we'd like this surface to be `visited' or `swiped' by a robot arm with fixed geometric degree fo freedom. In \cite{Atkar:2001wz}, Atkar et al. proposed a slice based method to decide a painting path for 3d surface. This method uses a topology method to decomposte the domain into convex pieces by the critical points in Reeb Graph. However, the generated path is not on the target domain but on a ``painting arm domain'' that is with fixed distance of the target domain, also, the path is not unique and depend on the angle of the slicing plane. 
%The other related work is called ``space filling curve''\cite{Goswami:2015wt}\cite{Ban:2013im}, the goal of this problem is to find a path that can fill up a given surface in the limit. 

\section{Conclusion} % (fold)
\label{sec:conclusion}
In this paper, a brand new surface parameterization, \emph{holomorphic quadratic differentials}, is adopted to perform the coverage path planning for general surfaces with complex topology. The natural coordinates of holomorphic quadratic differentials inherently induce non-intersecting trajectories on surfaces. This property inspires us to develope a robot coverage path planning algorithm. Moreover, holomorphic quadratic differentials intrinsically bring a regular number of surface decomposition. By converting the surface decomposition to its doubled dual graph, robots can travel on the whole surface according to the Euler cycle with great coverage.
% section conclusion (end)

% \begin{small}
\bibliographystyle{IEEEtran}
\bibliography{jie,jiepub,pathcover,yuylin}
% \end{small}

%\input{appendix}

\end{document}